\documentclass[journal,10pt]{IEEEtran}
\usepackage[utf8]{inputenc}

\usepackage{amsmath}
\usepackage{amssymb}
\usepackage{amsfonts}
\usepackage{cite}
\usepackage{float}

\newcommand{\Rmnum}[1]{\expandafter\@slowromancap\romannumeral #1@}

\usepackage{booktabs}
\usepackage{bbm}
\usepackage{mathrsfs}
\usepackage[caption=false]{subfig}

\usepackage{graphicx}
\usepackage{epsfig}
\usepackage[numbers,sort&compress]{natbib}
\usepackage{xcolor,soul}
\usepackage{color}
\usepackage{enumerate}
\usepackage{extarrows}
\usepackage[ruled,linesnumbered,lined]{algorithm2e}  
\usepackage{algorithmic}
\usepackage{tensor}
\usepackage[colorlinks,linkcolor=blue,anchorcolor=blue,citecolor=blue,urlcolor=black]{hyperref}
\usepackage{url}
\usepackage[normalem]{ulem} 

\usepackage{eso-pic}

\begin{document}
\AddToShipoutPictureFG*{
  \AtPageUpperLeft{
    \hspace{1.8cm}
    \raisebox{-1.2cm}{
      \small Published in IEEE Transactions on Cognitive Communications and Networking
    }
  }
}

\title{Neuromorphic Wireless Split Computing with Resonate-and-Fire Neurons}
\author{Dengyu Wu, \IEEEmembership{Member,~IEEE}, Jiechen Chen,  \IEEEmembership{Member,~IEEE},~ H. Vincent Poor,  \IEEEmembership{Life Fellow,~IEEE},\\Bipin Rajendran,~\IEEEmembership{Senior Member,~IEEE},~Osvaldo Simeone,~\IEEEmembership{Fellow,~IEEE}
\thanks{D. Wu and J. Chen are with the Department of Engineering, King’s College London, London, WC2R 2LS, UK (email:\{dengyu.wu, jiechen.chen\}@kcl.ac.uk). H. Vincent Poor is with the Department of Electrical and Computer Engineering, Princeton University, Princeton, NJ 08544 USA (e-mail:poor@princeton.edu). B. Rajendran and O. Simeone are with the Institute for Intelligent Networked Systems,
Northeastern University London, One Portsoken Street, London, E1 8PH,
UK (email: \{b.rajendran, o.simeone\}@nulondon.ac.uk).\\
This work was supported by the European Research Council (ERC) under the European Union’s Horizon Europe Programme (grant agreement No. 101198347), by an Open Fellowship of the EPSRC (EP/W024101/1), by the EPSRC project (EP/X011852/1), and by the U.S. National Science Foundation under Grant ECCS-2335876.
 }
\vspace*{-0.5cm}
}

\maketitle

\IEEEpeerreviewmaketitle
\newtheorem{definition}{\underline{Definition}}[section]
\newtheorem{fact}{Fact}
\newtheorem{assumption}{Assumption}
\newtheorem{theorem}{Theorem}
\newtheorem{lemma}{\underline{Lemma}}[section]
\newtheorem{proposition}{\underline{Proposition}}[section]
\newtheorem{corollary}[proposition]{\underline{Corollary}}
\newtheorem{example}{\underline{Example}}[section]
\newtheorem{remark}{\underline{Remark}}[section]
\newcommand{\mv}[1]{\mbox{\boldmath{$ #1 $}}}
\newcommand{\mb}[1]{\mathbb{#1}}
\newcommand{\Myfrac}[2]{\ensuremath{#1\mathord{\left/\right.\kern-\nulldelimiterspace}#2}}
\newcommand\Perms[2]{\tensor[^{#2}]P{_{#1}}}
\newcommand{\note}[1]{[\textcolor{red}{\textit{#1}}]}

\begin{abstract}
Neuromorphic computing offers an energy-efficient alternative to conventional deep learning accelerators, particularly for real-time processing of time-series data. However, many edge applications, such as wireless sensing and audio recognition, generate streaming signals with rich spectral features that are not effectively captured by conventional leaky integrate-and-fire (LIF) spiking neurons. This paper investigates a wireless split computing architecture that employs resonate-and-fire (RF) neurons with oscillatory dynamics to process time-domain signals directly, eliminating the need for costly spectral pre-processing. By resonating at tunable frequencies, RF neurons extract time-localized spectral features while maintaining low spiking activity. This temporal sparsity translates into significant savings in both computation and transmission energy. Assuming an OFDM-based analog wireless interface for spike transmission, we present a complete system design and evaluate its performance on audio classification and modulation classification tasks. Experimental results show that the proposed RF-SNN architecture achieves comparable accuracy to conventional LIF-SNNs and ANNs, while substantially reducing spike rates and total energy consumption during inference and communication.
\end{abstract}


\section{Introduction}
\subsection{Motivation}
Modern edge intelligence requires energy-efficient processing of streaming data such as radio-frequency and audio signals. Traditional deep neural network (DNN) accelerators, designed primarily for batch processing, are often ill-suited to continuous, always-on time-series tasks due to their high power consumption. In contrast, neuromorphic computing has emerged as a compelling alternative, shifting toward online, event-driven computation that more naturally aligns with the temporal nature of such data streams \cite{davies2021advancing, jang2019introduction, chen2023neuromorphic, 10606014}. 
\emph{Spiking neural networks} (SNNs), the computational model underpinning neuromorphic systems, process information through asynchronous binary events---\emph{spikes}---and inherently conserve energy by activating only in response to meaningful changes in the input signal. This \emph{sparse, event-driven} operation offers significant energy efficiency advantages over conventional continuous-valued networks \cite{ferreira2021neuromorphic, orchard2021efficient}.

Many signal processing tasks at the edge involve spectrally rich time-domain signals. Wireless baseband signals and audio waveforms, for example, encode critical information across frequency bands over time-features essential for applications such as modulation classification and speech recognition \cite{9311226, girmay2023technology}. However, conventional neuromorphic systems typically employ spiking neurons based on the \emph{leaky integrate-and-fire} (LIF) model, which integrates input over time but lacks inherent frequency selectivity. Consequently, SNNs often rely on pre-processing steps -- such as fast Fourier transforms (FFTs) or spectrograms -- to extract frequency-domain features prior to spike-based processing \cite{orchard2021efficient}. These features are optimized for signal reconstruction and not for the given inference task, resulting in high-dimensional inputs for LIF-based SNNs (see, e.g., \cite{bernardo2025symbol, 10447281}). Processing such inputs requires the deployment of sufficiently large models, potentially undermining the energy and latency benefits of neuromorphic inference, especially when deployed on constrained edge devices.

\emph{Resonate-and-fire} (RF) neurons offer a biologically inspired alternative better suited to such tasks. First introduced in \cite{izhikevich2001resonate}, RF neurons exhibit sub-threshold membrane potential oscillations and resonate at preferred frequencies. Functionally, they act as band-pass filters: they accumulate oscillatory input and fire spikes only when the frequency content of the input matches the neuron’s natural resonance. This enables in-neuron extraction of time-localized spectral features that are optimized for the given signal processing task.

While the RF neuron was originally introduced as a simplified model of biological behavior, in this work, we focus on its utility within an engineered decision-making system. Building on this principle, recent work has introduced \emph{balanced RF} (BRF) neuron models \cite{higuchi2024balanced}, which incorporate adaptive refractory mechanisms that control spiking activity while preserving the memory of past inputs. These enhanced RF models allow for extended temporal integration and exhibit sparse, frequency-selective firing-activating primarily when resonant input patterns are detected. 

\begin{figure*}[!t]
    \centering
    \includegraphics[width=0.9\linewidth]{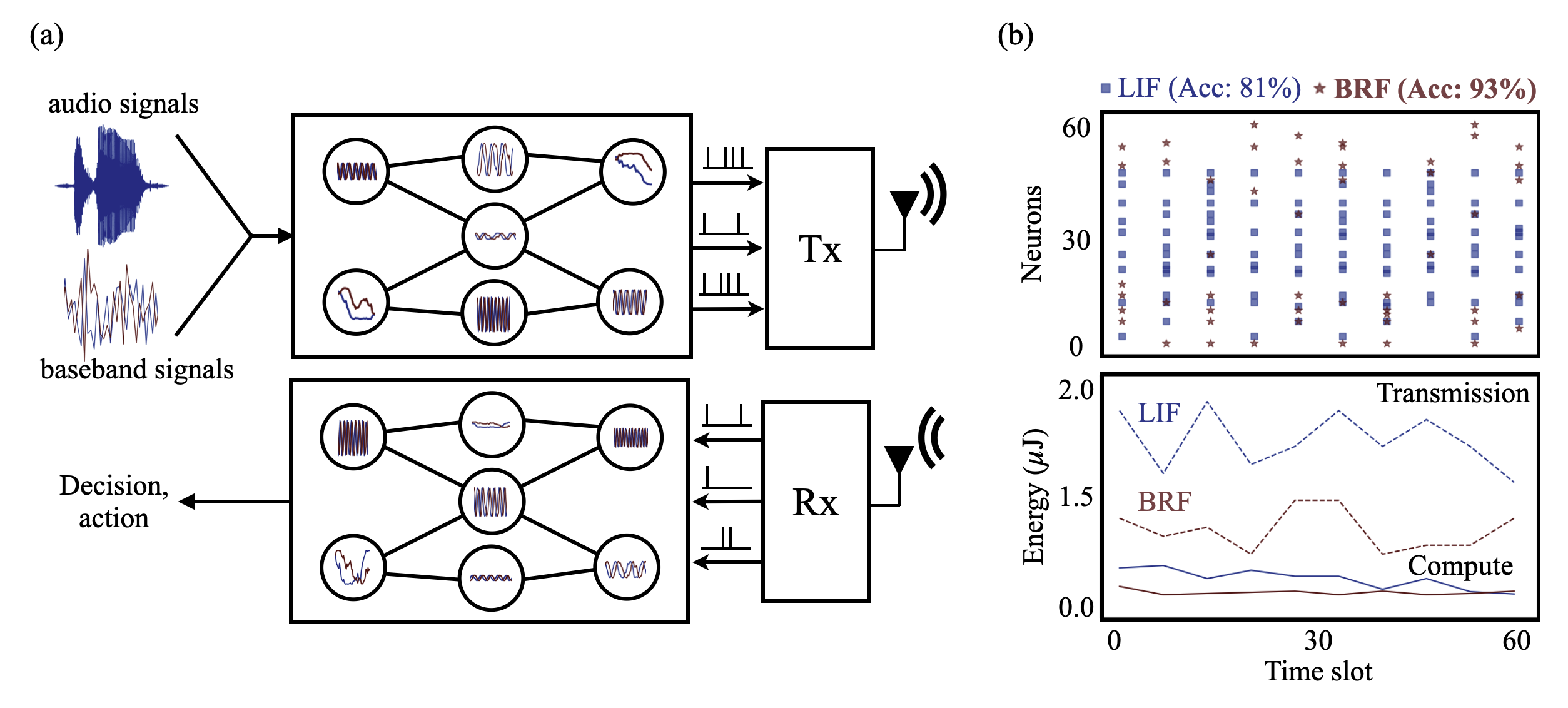}
    \caption{Neuromorphic wireless split computing architecture based on RF-SNNs: (a) The transmitter acquires time-domain signals with informative spectral components, such as audio or baseband (radio) signals. Upon processing by an SNN with balanced resonate-and-fire (BRF) neurons \cite{higuchi2024balanced}, the transmitter communicates with a receiver over a wireless channel using an OFDM interface. Following the split computing principle, a final decision or action is taken by the receiver based on an RF-SNN. (b) Results for audio classification \cite{higuchi2024balanced} (see Sec. \ref{sec:exp} for details): an SNN with BRF neurons tends to produce fewer spikes, while also improving accuracy (top). This, in turn, reduces transmission energy levels for the neuromorphic wireless split computing system (bottom).  }
    \label{fig:system}
\end{figure*}

By embedding a learned frequency analysis within the dynamics of individual neurons, RF and BRF neurons can directly process raw audio and radio-frequency signals, substantially reducing model size and spike rates compared to LIF neurons. In fact, as shown in Fig. \ref{fig:system}(b), BRF neurons fire only on resonance events, in contrast to the more frequent threshold crossings typical of LIF models.  Since neuromorphic systems consume energy primarily in proportion to the number of spikes, such sparsity directly translates to lower energy usage \cite{wu2022little, wu2024direct}.

Prior art focused on the use of the BRF neuron model in conventional centralized classification scenarios, in contrast, this paper explores the use of SNNs with BRF neurons -- RF-SNNs -- for wireless split computing. In wireless split computing, an SNN is partitioned across a transmitter and a receiver, with spike events transmitted over the air \cite{chen2023neuromorphic, wu2024neuromorphic, 10682971, chen2022neuromorphic}. In this architecture, the boundary between network partitions becomes a communication bottleneck: each spike generated at the transmitter must be conveyed to the receiver. As such, reducing spike activity not only lowers the computational energy required at each node but also significantly reduces the energy cost of wireless transmission.

By combining RF neurons’ spectral sensitivity with a low-overhead analog wireless interface, the proposed neuromorphic split computing architecture is uniquely suited to efficiently process spectrally rich time-series signals -- such as those encountered in wireless sensing and audio applications -- while minimizing both computation and communication energy. This paper presents a comprehensive system design as well as an end-to-end energy analysis encompassing both computation and wireless transmission. We evaluate wireless split on two distinct tasks, including audio and radio signal classification, thereby demonstrating the advantages of BRF neurons in a distributed neuromorphic system.

\subsection{Related Work}
\noindent \emph{Neuromorphic computing for time-series:} Neuromorphic hardware and SNN algorithms have been explored as a path to energy-efficient continuous sensing and inference. Unlike frame-based artificial neural networks, SNNs process information through discrete spikes and can inherently adapt their activity to the “semantics” of the data by consuming energy only when relevant events occur \cite{chen2023neuromorphic}. This makes them well-suited for streaming inputs.

Reference \cite{yin2021accurate} demonstrated that adaptive spiking neurons known as adaptive LIF (ALIF) models can achieve accurate and efficient classification of time-domain sequences, highlighting the importance of neuron models that capture temporal dependencies. First introduced by \cite{izhikevich2001resonate}, RF neurons were shown to produce spike outputs based on resonance phenomena rather than purely integrative thresholding. The Adaptive Exponential (AdEx) model \cite{brette2005adaptive} is also widely used in neuromorphic hardware, as it can exhibit resonance and bursting through adaptation currents. However, both the original RF neuron \cite{izhikevich2001resonate} and AdEx are computationally intensive, requiring the solution of coupled nonlinear differential equations. 

In order to obtain a more scalable and efficient computational model, reference \cite{higuchi2024balanced} recently proposed the BRF neurons, which enhance the basic RF model with tunable damping and refractory parameters. BRF neurons exhibit prolonged memory and sparse firing thanks to their resonance-driven accumulators and adaptive thresholds. This makes them especially suitable for tasks requiring extraction of temporal-frequency features, as they can be configured to emphasize desired frequency responses while suppressing spurious spikes.

Reference \cite{shrestha2024efficient} applied resonate-and-fire neurons on Loihi 2 for standard video and audio processing tasks, reporting 400$\times$ 
improvement in the energy–latency product compared to state-of-the-art DNN approaches. 
These results underscore the fact that, when implemented on appropriate hardware, oscillatory spiking dynamics can yield massive efficiency gains for processing spectral or temporal data. Overall, the literature suggests that leveraging complex spiking neuron models (adaptive or resonant) and neuromorphic hardware can drastically reduce power consumption for continuous sensory data processing \cite{yang2022lead}, although this requires careful optimization of the network model to preserve task accuracy.

\noindent \emph{Wireless neuromorphic communication:} There is a growing body of work integrating SNNs with wireless communication and edge computing systems. In \emph{split neuromorphic computing}, an SNN model is divided between a transmitter (edge device) and a receiver (edge server or access point), and hence intermediate features must be sent over the network. Reference \cite{chen2023neuromorphic} presented a design for neuromorphic wireless cognition, in which each edge device uses spike-based sensors and SNN encoders to transmit semantic information over a fading channel to an SNN-based receiver. The entire system—spike generation, wireless transmission, and SNN decoding—was jointly trained, and it achieved significant reductions in latency and energy consumption over conventional frame-based methods.
 
Reference \cite{wu2024neuromorphic} recently studied a wireless split SNN architecture and explored the use of \emph{multi-level spikes} to improve inference accuracy. By allowing each spike to carry a few bits of analog amplitude information (instead of being strictly binary), they observed performance gains at the cost of higher communication requirements. This work included both digital and analog transmission strategies for the multi-level spike streams, demonstrating the viability of directly sending neuromorphic signals over standard wireless channels.

The concept of transmitting graded or analog-valued spikes ties into broader efforts on analog neuromorphic communication. For example, reference \cite{ferreira2021neuromorphic} has shown that analog spiking modulators can encode analog signals with extreme energy efficiency (on the order of femtojoules per spike conversion) using silicon neuron circuits. These advances reinforce the idea that spike-based information can be communicated with minimal energy, especially if the number of spikes is kept low.

In a related line of work, reference \cite{yang2022lead} proposed a federated learning framework using neuromorphic principles for wireless edge machine learning. By training SNN models collaboratively across devices in a federated setting, this work reduces uplink data traffic compared to standard federated learning, with only a minor accuracy trade-off.

\noindent \emph{Neuromorphic edge inference and applications:} Beyond communications, integrating neuromorphic computing at the network edge has shown promise in a variety of sensing tasks. Of particular interest for this work are efforts that have applied SNNs to on-device audio and sensor data processing with custom hardware. Reference \cite{shrestha2024efficient}, mentioned earlier, demonstrated that neuromorphic processors (Loihi 2) can handle audio and video streams with far superior energy-delay efficiency than conventional methods. These results are encouraging for applications like always-on audio classification or real-time spectrum sensing in wireless systems, which demand low power consumption.

To facilitate research in these areas, benchmark datasets and tasks have also been established. The \emph{Spiking Heidelberg Digits (SHD)} dataset introduced by \cite{9311226} provides a standard suite of event-based auditory signals (spoken digits) for evaluating SNN temporal classification. Likewise, in the wireless domain, reference \cite{girmay2023technology} compiled an \emph{ITS radio signal} dataset for technology modulation recognition, consisting of complex baseband waveforms from different wireless standards.

In summary, the literature spans from neuron-level model innovations \cite{izhikevich2001resonate, higuchi2024balanced, yin2021accurate} to system-level wireless SNN implementations \cite{chen2023neuromorphic, 10682971, wu2024neuromorphic}, all pointing toward the potential of neuromorphic computing to enable low-energy, low-latency processing of streaming data in various edge scenarios. Our work extends this line by uniting oscillatory spiking neurons with a wireless split computing architecture and quantifying the resulting benefits in both accuracy and energy.

\subsection{Main Contributions}
In this paper, we propose a novel neuromorphic wireless split computing architecture utilizing BRF spiking neurons with the aim of supporting the efficient processing of spectrally rich radio or audio signals. The main contributions are as follows:
\begin{itemize}
\item \textbf{RF neurons for wireless neuromorphic split computing:} We propose the use of RF spiking neurons in a split SNN to directly process raw time-domain signals with rich spectral content. RF neurons inherently act as band-pass filters, enabling the distributed system to extract and leverage task-specific frequency-domain features. This yields lower-dimensional features with significantly decreased spiking activity compared to conventional LIF-based SNNs with or without conventional pre-processing steps such as FFTs.

\item \textbf{System design and energy consumption analysis:} As illustrated in Fig. \ref{fig:system}, we develop a split SNN architecture in which an encoding SNN at the transmitter and a decoding SNN at the receiver communicate via spike events over a wireless link. The system is implemented assuming an OFDM-based analog radio interface in which spikes are modulated onto OFDM subcarriers. We analytically characterize the overall energy consumption, including both computation and communication, incorporating a regularization mechanism during training to promote spike sparsity and reduced energy requirements.

\item \textbf{Experimental evaluation on audio and radio-frequency classification tasks:} We evaluate the proposed RF-SNN split computing system on two representative tasks: (\textit{i}) neuromorphic audio classification, using the SHD spiking auditory dataset \cite{9311226}, and (\textit{ii}) RF modulation classification, using a wireless signal dataset in the ITS band \cite{girmay2023technology}. Note that the first dataset consists of spiking signals produced by a neuromorphic front-end, while the second includes natural signals. We compare three alternative solutions: an SNN with LIF neurons \cite{wu2024neuromorphic}, an SNN with BRF neurons \cite{higuchi2024balanced}, and a conventional artificial neural network (ANN) baseline \cite{girmay2023technology}. The results show that the BRF-SNN achieves competitive accuracy, while reducing the spike count, and thus the energy consumption, by a substantial margin.
\end{itemize}

\section{Resonate-and-Fire Spiking Neural Network} \label{MED}
As illustrated in Fig.~\ref{fig:system}(a), this paper studies a neuromorphic wireless split computing system that is designed to process input with strong spectral features, such as audio or radio-frequency/baseband signals. An application is modulation classification, in which radio signals collected at the transmitter side are collaboratively processed by two nodes connected over a wireless link, enabling the detection of the modulation scheme at the receiver side. Another application is speech recognition for a personal assistant, in which speech captured by a microphone at the transmitter is decoded into text at the receiver through a model split between the transmitter and receiver.

In order to efficiently process informative features in the frequency domain, we investigate the use of neuromorphic computing platforms based on spiking neuronal models with oscillatory dynamics, which are known as \textit{resonate-and-fire (RF) neurons} \cite{izhikevich2001resonate, higuchi2024balanced}. This setting differs from the prior art on neuromorphic wireless split computing that adopts instead conventional LIF neurons \cite{chen2023neuromorphic, wu2024neuromorphic}. As exemplified in Fig.~\ref{fig:system}(b), the use of  RF spiking neurons will be shown to result in a higher degree of sparsity in the number of spikes produced by the network, yielding  higher energy efficiency while also improving the accuracy. 

In this section, we first describe the considered SNN model based on \emph{balanced RF (BRF) neuron} model proposed in \cite{higuchi2024balanced}. BRF neurons exhibit long memory and sparse firing rates due to resonance-driven activity and adaptive refractory mechanisms, making them uniquely suited for the processing of time series in communication, radar, and audio processing. We then describe the split neural network architecture that will be designed and analyzed in the next section.

\subsection{Balanced Resonate-and-Fire (BRF) Neurons}
The RF neuron, introduced by Izhikevich \cite{izhikevich2001resonate}, models neurons with subthreshold oscillations that exhibit a resonance-dependent spiking mechanism. This paper adopts BRF neurons, which provide a more flexible model to control the interplay between memory and spiking rate \cite{higuchi2024balanced}. 

To explain this model, consider a layered architecture, with the pair of indices $i$ and $l$ denoting neuron $i$ in layer $l$. All BRF neurons in the network share a common update rate parameter $\delta>0$, and each neuron $i$ in layer $l$ has a local angular frequency $w_i^l$, as well as a time-varying damping factor $b_{t,i}^l$, where $t$ is the discrete time index. The complex membrane potential of BRF neuron $i$ in layer $l$, denoted as $u_{t,i}^l \in \mathbb{C}$, evolves according to the resonator dynamics \cite{higuchi2024balanced}
\begin{equation}\label{eq:rf_dynamics}
   u_{t,i}^l=u_{t-1,i}^l+\delta((b^l_{t,i}+j\omega^l_i)u_{t-1, i}^l+I_{t,i}^l), 
\end{equation}
where $I_{t,i}^l$ is a generally complex-valued input at time $t$.

The evolution in \eqref{eq:rf_dynamics} defines the membrane potential $u_{t,i}^l$ of a BRF neuron as a damped sinusoidal oscillation driven by input current, where parameter $\omega_i^l$ sets the oscillation frequency. For stability, the damping factor must satisfy the inequality $\delta\omega^l_i < 1$, which ensures that the oscillatory dynamics decay over time. By \eqref{eq:rf_dynamics}, BRF neurons implement damped resonator dynamics that are mathematically equivalent to an exponentially windowed bandpass filter, analogous to individual bins of a short-time Fourier transform \cite{orchard2021efficient}.

The output of neuron $i$ in layer $l$ is given by 
\begin{equation}
S_{t, i}^l=\left\{
\begin{array}{ll}
0~\text{(no spike)},        & \text{if}~ \operatorname{Re}(u_{t, i}^l) \leq \vartheta_{t,i}^l, \\
1~\text{(spike)},       &   \text{if}~ \operatorname{Re}(u_{t, i}^l) > \vartheta_{t,i}^l,
\end{array} \right. \label{eq:spike}
\end{equation}
where $\operatorname{Re}(u_{t,i}^l)$ represents the real part of the membrane potential $u_{t,i}^l$, and $\vartheta_{t,i}^l$ is a  threshold. The neuron outputs a spike when its membrane voltage crosses the threshold; the real part is taken as the effective voltage, while the imaginary part captures oscillatory dynamics without directly triggering spikes.

The BRF neuron controls the spiking rate via a \textit{refractory period} mechanism based on the dynamic modulation of the threshold $ \vartheta_{t,i}^l$. Specifically, the threshold is updated as
\begin{equation}\label{eq:adap_vthr_brf}
    \vartheta_{t,i}^l = \vartheta^l_c + q_{t,i}^l, 
    \quad \text{where} \quad q_{t,i}^l = \gamma^l q_{t-1,i}^l + S_{t - 1,i}^l,
\end{equation}
where $ \vartheta^l_c$ is a baseline threshold, and $\gamma^l \in [0,1)$ is a hyperparameter determining the duration of the refractory period. By \eqref{eq:adap_vthr_brf}, if the neuron spikes at the previous time $t-1$, i.e., if $S_{t - 1,i}^l=1$, the variable $q_{t,i}^l $ is increased. This, in turn, increases the threshold $\vartheta_{t,i}^l$ in \eqref{eq:adap_vthr_brf},  making it more unlikely that the neuron spikes at time $t$. The parameter $\gamma^l$ determines the duration of the refractory period, with a larger value of $\gamma^l$ producing a longer refractory period.

To enhance stability, the BRF neuron also relates the dampening factor $ b_{t,i}^l$ in \eqref{eq:rf_dynamics} to the cumulative spiking rate $q_{t,i}^l$ in \eqref{eq:adap_vthr_brf} and to the angular frequency $w_i^l$ as
\begin{equation} \label{eq:dampening}
    b_{t,i}^l = \frac{-1 + \sqrt{1 - (\delta \omega^l_i)^2}}{\delta} - \hat{b}_{i}^{l} - q_{t,i}^l,
\end{equation}
where $ \hat{b}_{i}^{l} > 0 $ is a trainable offset parameter. Due to \eqref{eq:dampening}, a spike at the previous time, i.e., $S_{t - 1,i}^l=1$, yields a faster decrease of the membrane potential, which further reduces the spike rate.

\section{Resonate-And-Fire Neuromorphic wireless split computing}
\begin{figure}[!t]
    \centering
    \includegraphics[width=0.95\linewidth]{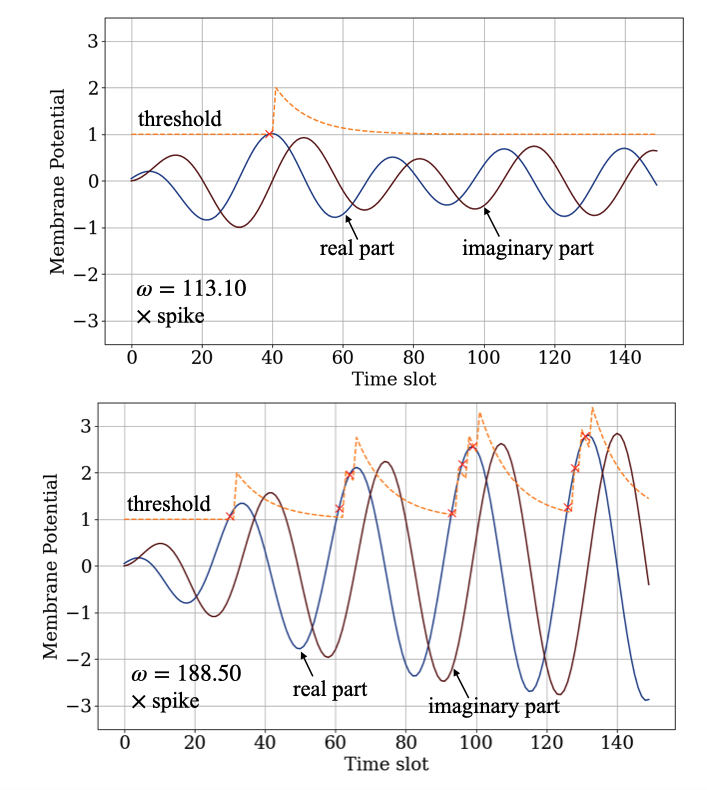}
    \caption{Illustration of the dynamic of the BRF neuron ($\hat b=15$) when the input is a sinusoid with frequency $188.50~\rm{rad/s}$ \cite{higuchi2024balanced}: A spike is generated when the real part of the membrane potential exceeds a time-varying threshold (dashed line), which adapts to the spiking activity of the neuron. The top plot corresponds to a BRF neuron with intrinsic frequency $\omega = 113.10~\rm{rad/s}$, which does not match the frequency of the input, while the bottom plot corresponds to a BRF neuron with intrinsic frequency $\omega = 188.50~\rm{rad/s}$, which aligns with that of the input.}
    \label{fig:wireless_cn}
\end{figure}

As illustrated in Fig.~\ref{fig:system}(a), we consider a neuromorphic wireless split computing architecture for remote inference. Targeting the processing of signals with informative frequency-domain features, we study a system in which an SNN  split between transmitter (Tx) and receiver (Rx) implements BRF neurons.  

Unlike previous works \cite{chen2023neuromorphic, wu2024neuromorphic} which utilize conventional LIF neurons, BRF neurons are adopted, as shown in Fig.~\ref{fig:wireless_cn}, since they are particularly well-suited to extract frequency-domain features from inputs such as audio signals and baseband signals in wireless communications. For such signals, BRF neurons can enhance performance while also reducing the spike rate \cite{shrestha2024efficient}. This creates a unique opportunity for split computing architectures, in which smaller spike rates directly translate into lower transmission power \cite{chen2023neuromorphic}.

\subsection{Split SNN}
Consider a layered pre-trained SNN encompassing neurons that follow one of the models described in the previous section. The SNN is trained as detailed in Section \ref{sec:regularizer_app} by targeting a regularized learning objective that aims at maximizing training accuracy while penalizing large spike rates.

As shown in Fig.~\ref{fig:system}, the pre-trained SNN is split across a given layer between an encoding SNN deployed at the Tx and a decoding SNN implemented at the Rx. As detailed next, the spike signals produced by the last layer of the encoding SNN are communicated to the decoding SNN over a wireless channel, which is subject to fading and noise.

\subsection{Transmit-Side Processing}
\begin{figure}[ht!]
    \centering
    \includegraphics[width=0.95\linewidth]{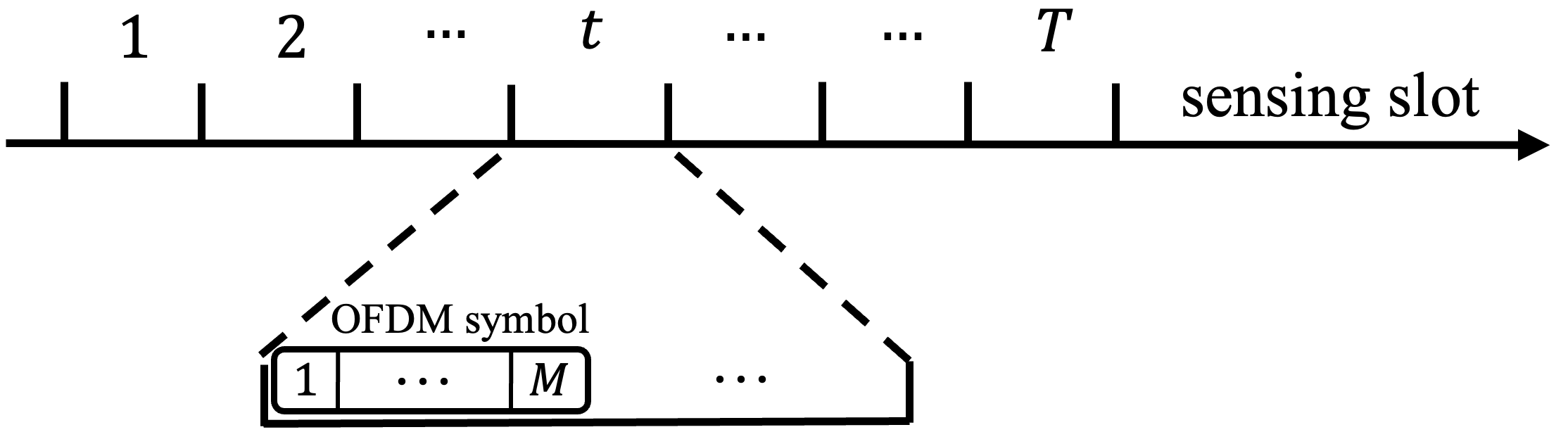}
    \caption{Timeline of a neuromorphic wireless split computing system \cite{wu2024neuromorphic, chen2023neuromorphic, 10682971}. Time is discretized into sensing slots $t=1,2,\ldots,T$. During each sensing slot, a sample of the sensed signal is acquired and processed by the encoding SNN. The spike signals produced by the output layer of the encoding SNN are encoded using OFDM. As illustrated in the figure, the duration of a sensing slot is typically larger than that of the OFDM symbol, e.g., a sensing slot of 4 ms for an audio signal and an OFDM symbol of duration 35.68 $\mu \rm s$ \cite{NR}. This means that one sensing slot is always sufficient to transmit the spikes generated in the previous slot, so the OFDM transmission does not introduce additional latency. }
    \label{fig: timeline}
\end{figure}

As illustrated in Fig.~\ref{fig: timeline}, the neuromorphic split computing system operates in discrete time $t=1,\ldots,T$ over $T$ sensing slots. During the $T$ sensing slots, the Rx collects information from the Tx, and a decision is made at the Rx at the end of the $T$-th sensing slot. The Tx and Rx are both equipped with a single antenna. 

At each slot $t=1,\ldots, T$, the SNN at the Tx processes the incoming signals $\mv X_t=[X_{t,1}, \ldots, X_{t,D}]^T$ producing the spike signals $\mv S_t=[S_{t,1}, \ldots, S_{t,M}]\in\{0,1\}^M$ at its last layer. The input signals $X_{t,i}$ may be complex-valued, as is the case for baseband signals. The spike signals $\mv S_t$ are modulated into an OFDM symbol $\mv x_t$ and transmitted to the Rx.

Each OFDM symbol consists of a set  $\mathcal{N}=\mathcal{N}^{\rm D} \cup \mathcal{N}^{\rm P}$ of subcarriers, with $|\mathcal{N}^{\rm D}|= N^{\rm D}$ data subcarriers and $|\mathcal{N}^{\rm P}|=N^{\rm P}$ pilot subcarriers. Each data subcarrier is assigned to modulate the spike generated by a given neuron. Accordingly, the number of data subcarriers equals the number of output neurons in the encoding SNN, i.e., $N^{\rm D}=M$.  Furthermore, the $m$-th subcarrier modulates the signal 
\begin{align}
    x_{t,m}=\left\{
\begin{array}{ll}
\sqrt{P}S_{t,m},       &\text{if}~ m \in \mathcal{N}^{\rm D}, \\
x^{\rm p},         & \text{if}~ m \in \mathcal{N}^{\rm P}.
\end{array} \right. \label{eq: conventional_spike}
\end{align}
where $P$ is the per-subcarrier transmission power, and $x^{\rm p}$ is the pilot signal with power $|x^{\rm p}|^2=P$. More generally, one may consider settings in which multiple subcarriers are assigned to each output neuron of the encoding SNN, and/or multiple OFDM symbols are transmitted for each sensing slot. 

The overall transmission energy at slot $t$ is given by
\begin{equation}
    E_{\text{tx}, t}= \sum_{m=1}^M S_{t,m} P, \label{tran_ene}
\end{equation}
which is thus proportional to the number of spikes, $\sum_{m=1}^MS_{t,m}$, modulated by the OFDM symbol in \eqref{eq: conventional_spike}.

\subsection{Receive-Side Processing}
Assuming each OFDM symbol’s cyclic prefix is longer than the multi-path channel’s delay spread, the received OFDM symbol in sensing slot $t$ is given by \cite{goldsmith2005wireless}
\begin{align}
    \mv y_t= \mv H_t \mv x_t + \mv w_t, \label{receiving}
\end{align}
where $\mv H_t$ is a diagonal channel matrix containing the channel frequency responses across all subcarriers, and $\mv w_t$ is a noise vector with i.i.d. complex Gaussian elements of zero mean and variance $N_0$.

The receiver processes the received signal $\mv y_t$ by estimating the channel $\mv H_t$, and then performing equalization. The channels on the pilot subcarriers are estimated as $\hat{h}_{t,m}^{\rm p}=y_{t,m}/x^{\rm p}$, for all $m\in\mathcal{N}^{P}$. The channels $h_{t,m}^{\rm D}$ on the data subcarriers are then estimated using interpolation based on the estimates $\hat{h}_{t,m}^{\rm p}$ on the pilot subcarriers \cite{goldsmith2005wireless}. Then the $m$-th received data symbol is equalized by treating the estimated channel as real, yielding the equalized symbol as $\hat{x}_{t,m}=y_{t,m}/\hat{h}_{t,m}^{\rm D}$. Finally, the equalized symbol $\hat{x}_{t,m}$ is mapped to an estimated spiking signal $\hat{S}_{t,m}$ using a threshold-based rule applied to its real part as
\begin{align}
    \hat{S}_{t,m}=\left\{
\begin{array}{ll}
1,       &\text{if}~ \textrm{Re}(\hat{x}_{t,m})>1/2, \\
0,         & \text{otherwise},
\end{array} \right. 
\end{align}
The collection of spiking signal $\hat{\mv S}_t=[\hat{S}_{t,1}, \ldots, \hat{S}_{t,M}]$ is fed to the decoding SNN.

OFDM modulation, demodulation, pilot processing, and channel interpolation are standard operations in the transceiver chain of a wireless device. Their cost is identical across all neuron models and independent of the spiking dynamics. For instance, a standard OFDM baseband receiver has been reported to consume about 98 mW using 0.18 $\mu m$ CMOS technology \cite{yu2012design}. This corresponds to roughly 3.5 $\mu \rm J$ per symbol for our symbol duration of 35.68 $\mu \rm s$. In contrast, the computing and communication energies in our system range from about 1–2 $\mu \rm J$ at short distances to 5–10 $\mu \rm J$ at longer distances (see Fig. 11). Since this represents a constant overhead across all compared cases, we omit it from the energy analysis, and focus on the differences due to neuron updates and spike transmission.

\subsection{Optimization}\label{sec:regularizer_app}
The operation of the outlined split computing system depends on the synaptic weights and on the internal parameters of the neuron in the overall SNN spanning the encoding and decoding SNNs. Denote the tensor of all the parameters by $\Theta$, which includes the input synaptic weights $w^l_{i,j}$, the local angular frequency $\omega^l_i$, and the offset parameter $\hat{b}^l_i$, for all neurons $i$.

\subsubsection{Training Loss}
To optimize the parameters $\Theta$, we adopt an objective that includes training loss and a sparsity-promoting regularizer. Focusing on $K$-way classification, we convert the output layer of the decoding SNN to a decision layer with a number of neurons equal to the number of classes $K$. The decision layer produces a probability $p_k$ for each class $ k = 1, \ldots, K$. For example, the decision layer may use the logits of a softmax function on the rates of different neurons in the output layer of the decoding SNN.

Given a dataset $ \mathcal{D}$, denoting as $ p_{d,t}(\Theta)$ the probability assigned to the true class for data point $ d \in \mathcal{D}$ at time $t $, the cross-entropy loss is given by \cite{wu2024neuromorphic}
\begin{equation}
\mathcal{L}_{\text{CE}, t}(\Theta) = -\sum_{d \in \mathcal{D}} \log p_{d,t}(\Theta)  
\label{eq:cross_entropy}
\end{equation}
Note that the probability $p_{d,t}(\Theta)$ depends on the parameters $\Theta$ through the spiking outputs of the neurons in the decoding SNN.

\subsubsection{Sparsity-promoting Regularizer}
We also introduce a regularizer to regulate the sparsity of the spike signals produced by the neurons of the encoding and decoding SNNs. The introduction of this term aims at controlling the energy consumption of the system. In fact, as described in Section \ref{energy}, the compute energy consumption depends on the number of spikes produced by the SNNs (see \eqref{eq:energy_soma}), and the transmission energy is proportional to the number of output spikes (see \eqref{tran_ene}).  

From \eqref{eq:spike}, a spike is produced when the real part of the membrane potential $u^l_{t,i}$ of neuron $i$ in layer $l$ exceeds its spiking threshold $\vartheta_{t,i}^l$. Thus, to lower the spiking activity, one can decrease the absolute value of the membrane potential. In a manner similar to \cite{hoyer2004non} and \cite{Yang2020DeepHoyer}, we quantify the level of sparsity in the normalized membrane potentials at each layer $l$ by calculating the ratio between its $\ell_1$-norm and $\ell_2$-norm. For layer $l$ and time $t$, this yields the regularization term
\begin{equation} \label{eq:regularizer}
    \mathcal{R}_t^l(\Theta) = \frac{\left( \sum_i |\hat u^l_{t,i}| \right)^2}{\sum_i |\hat u_{t,i}^l|^2}, \quad \text{where} \quad
    \hat u^l_{t,i} = \max \left( \frac{u^{l}_{t,i}}{\vartheta_{t,i}^l },0 \right).
\end{equation}
The regularizer \eqref{eq:regularizer} is scale-invariant and promotes sparsity by encouraging fewer entries of vector $\boldsymbol{u}^l_t=[u^l_{t,1}, \ldots, u^l_{t,K^l}]^T$ to be larger than zero, i.e., $K^l$ is the number of neurons in layer $l$.

\subsubsection{Training objective} 
Overall, the training objective is given by
\begin{equation} \label{eq:l_reg}
    \mathcal{L}(\Theta) = \frac{1}{T} \sum_{t=1}^T \left( \mathcal{L}_{\rm{CE}, t}(\Theta) + \alpha \sum_{l} \mathcal{R}_t^l(\Theta)  \right),
\end{equation}
where hyperparameter $\alpha$ determines the trade-off between accuracy and spike sparsity.

\subsection{Weight Quantization and Calibration} \label{sec:quantization}

The encoding and/or decoding SNNs are intended for deployment on resource-constrained devices. Therefore, after training, the model parameters must be practically quantized to fit the memory and computational constraints of the devices \cite{davies2018loihi}. To this end, we apply quantization-aware fine-tuning based on calibration using the straight-through gradient estimator \cite{bengio2013estimating}. 

For each synaptic weight $w_{i,j}^l$ in layer $l$, we adopt a uniform quantizer
\begin{equation}
Q^l(w_{i,j}^l) = \frac{ \left\lfloor \lambda ^l w_{i,j}^l \right\rceil}{\lambda ^l}, \quad \text{where} \quad \lambda ^l = \frac{2^{m-1}}{\displaystyle\max_{i,j}\left| w_{i,j}^l\right|},
\end{equation}
with $\left\lfloor \cdot \right\rceil$ denoting the nearest-integer rounding operation, and $m$ is the bit precision. 

To mitigate accuracy degradation from direct quantization of a pre-trained SNN model, we adopt layer-wise calibration~\cite{hubara2021accurate}. Calibration attempts to reduce the discrepancy between the spikes $\hat{S}_{t,i}^{l}$ produced by the $K^l$ neurons at layer $l$ in the quantized model and the original spikes $S_{t,i}^{l}$ of the full-precision model. This is done by minimizing the loss function
\begin{equation}
\mathcal{L}^l_{\rm{cal},t}(\Theta) = \frac{1}{K^l} \sum_{i=1}^{K^l} \left( \hat{S}_{t,i}^{l} - S_{t,i}^{l} \right)^2,
\label{eq:calibration}
\end{equation}
averaged over layers $l$ and time instant $t$. This optimization targets neuron parameters $\Theta$, including local angular frequency $\omega^l_{i}$ and offset parameter $\hat b^l_{i}$.

\section{Modeling Computation Energy Consumption}  \label{energy}
In this section, we introduce a model for the energy consumption of the computing elements of the proposed neuromorphic split computing architecture. For reference, we also consider a baseline implementation based on LIF neurons \cite{yin2021accurate}. 

\subsection{ALIF vs. BRF neurons}
Current SNN implementations are typically based on LIF neurons and variants thereof. LIF neurons operate on real-value inputs, which are low-pass filtered prior to thresholding to produce output spikes. Thus, LIF neurons lack the capacity to efficiently capture complex dynamics in the frequency domain. The \textit{Adaptive LIF} (ALIF) neuron introduced in \cite{yin2021accurate} enhances the standard LIF model by incorporating a dynamic firing threshold that adapts based on recent spiking history.  Given a real-valued input current $I_{t,i}^l$, the real-valued membrane potential $u_{t, i}^l$ of the ALIF neuron $i$ in layer $l$ evolves as 
\begin{equation} \label{eq:alif_dynamics}
    u_{t, i}^l = \sigma(\hat{b}_{i}^{l})u_{t-1, i}^l + (1-\sigma(\hat{b}_{i}^{l})) I_{t,i}^l - \underbrace{S_{t - 1,i}^l \vartheta_{t-1,i}^l}_{\text{soft reset}} 
\end{equation}
where $\sigma(\hat{b}_{i}^{l}) = \exp(-1/\hat{b}_{i}^{l})$ is an exponential decay factor parametrized by a trainable parameter $\hat{b}_{i}^{l} \in \mathbb{R}^+$. A soft reset mechanism is applied by subtracting the adaptive threshold $ \vartheta_{t-1,i}^l$ from the membrane potential $u_{t, i}^l$ after a spike at the previous time $t-1$. The spiking output $S_{t,i}^l$ of the ALIF neuron is then given by 
\begin{equation}
S_{t, i}^l=\left\{
\begin{array}{ll}
0~\text{(no spike)},        & \text{if}~ u_{t, i}^l \leq \vartheta_{t,i}^l, \\
1~\text{(spike)},       &   \text{if}~u_{t, i}^l > \vartheta_{t,i}^l.
\end{array} \right. \label{eq:spike_alif}
\end{equation}

The firing threshold is updated adaptively based on recent spiking activity as
\begin{equation}\label{eq:adap_vthr_alif}
    \vartheta_{t,i}^l  = \vartheta_c^l + q_{t,i}^l, \quad \text{with} \quad 
    q_{t,i}^l = \beta \gamma^l q_{t-1,i}^l + \beta (1-\gamma^l)S_{t - 1,i}^l, 
\end{equation}
where $\vartheta_{c}^l$ is the baseline threshold, $\beta$ is a constant controlling the adaptation strength, and $\gamma^l\in(0,1)$ governs the decay rate of the adaptation variable $q_{t,i}^l$. Through the adaptive thresholding mechanism \eqref{eq:adap_vthr_alif},  ALIF neurons can control their firing rates. In this regard, note that the BRF neuron introduces a more complex dynamic, where the adaptation variable $q_{t,i}^l$ not only increases the threshold $\vartheta_{t,i}^l$ as in \eqref{eq:adap_vthr_brf}, but it also controls the damping factor $b_{t,i}^l$ of the oscillating membrane potential in \eqref{eq:rf_dynamics}.

\subsection{Energy Consumption Model}
Previous works commonly estimate the energy consumption of SNNs primarily based on the number of synaptic operations, which are often modeled as weight accumulation \cite{wu2022little,yang2022lead, wu2024direct, wu2024neuromorphic}. While this provides a reasonable approximation for conventional LIF-based SNNs, such estimates tend to overlook the computational overhead introduced by the neuronal (soma) dynamics. The contribution of somatic processing to the overall energy consumption of an SNN is particularly relevant when comparing architectures with different neuron models. In order to enable a fair comparison between SNNs based on ALIF and BRF neurons, in this section, we explicitly account for the energy contribution of both synaptic and soma operations. Ultimately, the energy consumption of different SNN models will be seen to depend on both the sparsity of the spiking activity and the per-neuron computational cost for the somatic dynamics. We do not include memory access energy in this analysis, focusing instead on the computational energy of different neuron models. Although spike patterns can affect memory access frequency, we assume these costs are minimized by hardware-level optimizations such as local memory access in Intel's Loihi \cite{davies2018loihi}.

\begin{figure}[ht]
    \centering
    \includegraphics[width=1.0\linewidth]{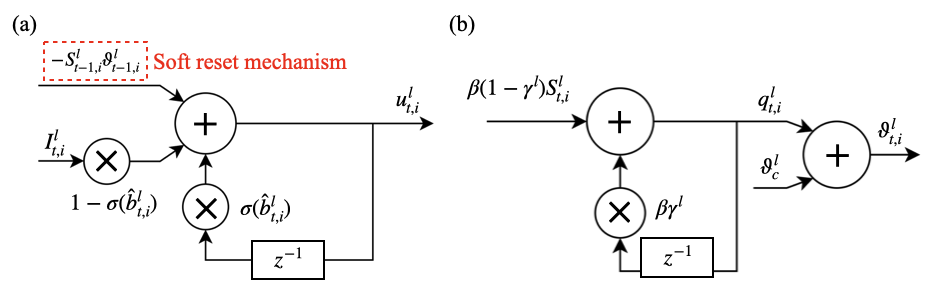}
    \caption{Digital implementation schematic of the somatic operation of an ALIF neuron: 
    (a) ALIF neuron dynamics \eqref{eq:alif_dynamics} and
    (b) adaptive threshold mechanism \eqref{eq:adap_vthr_alif}. 
    }
    \label{fig:digital_alif}
\end{figure}

\subsubsection{\textit{ALIF Neuron}}
A digital implementation schematic of the somatic operations for ALIF neurons can be found in Fig.~\ref{fig:digital_alif}. Implementing the updates \eqref{eq:alif_dynamics}, the membrane potential of the ALIF neuron is evaluated via the block diagram in Fig~\ref{fig:digital_alif}(a), requiring two multiplications and two additions. Note that the factors $1-\sigma(\hat{b}_{i}^{l})$ and $\sigma(\hat{b}_{i}^{l})$ involved in the products can be pre-computed and stored in the memory. The threshold update follows \eqref{eq:adap_vthr_alif}, which can be realized via the block diagram in Fig.~\ref{fig:digital_alif}(b). This requires one multiplier and two adders at each time step, with one additional multiplication and addition applied when a spike is generated. Overall, the somatic operation at each ALIF neuron amount to $N_{\mathrm{add}}^{\rm som}=2$ sums and $N_{\rm mul}^{\rm som}=3$ multiplications applied at each step, along with $N_{\mathrm{add}}^p=2$ sums and no multiplications, i.e., $N_{\rm mul}^p=0$, for post-spike processing.

The input current $I^l_{t,i}$ is computed through synaptic operations, which amount to one sum for each input spike. Accordingly, the number of synaptic sums equals the number of input spikes $\check M_{t,i}^l \leq n_i^l$, where $n_i^l$ is the number of pre-synaptic neurons, i.e., $N_{\rm add}^\textrm{syn} = \check M_{t,i}^l$ for each neuron $i$.

\begin{figure}[ht]
    \centering
    \includegraphics[width=1.0\linewidth]{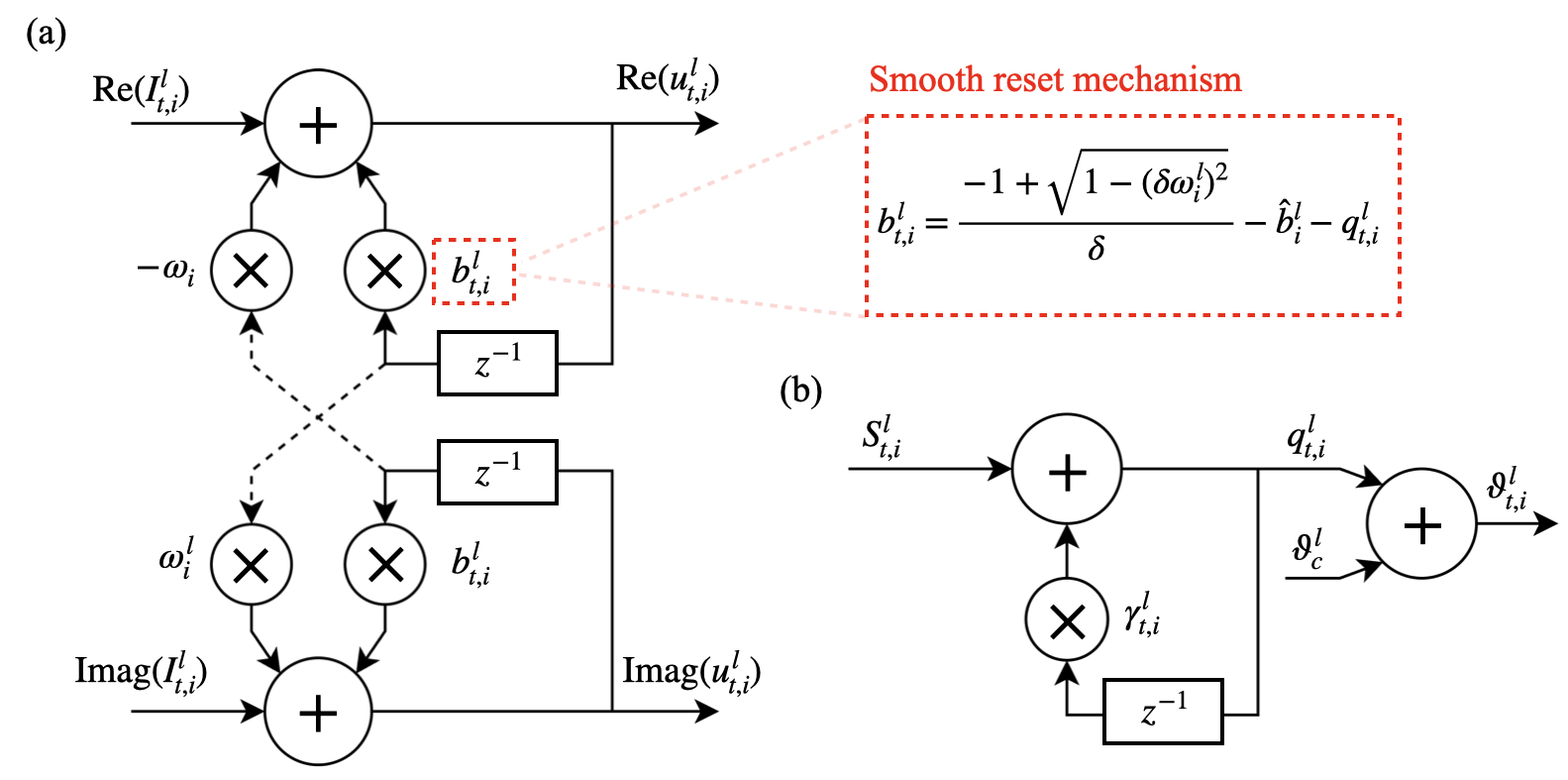}
    \caption{Digital implementation schematic of the somatic operation of a BRF neuron: 
    (a) BRF neuron dynamics \eqref{eq:rf_dynamics} with dampening factor $b_{t,i}^l$ in \eqref{eq:dampening}. 
    (b) Adaptive threshold mechanism \eqref{eq:adap_vthr_brf}.  
    }
    \label{fig:digital_brf}
\end{figure}

\subsubsection{\textit{BRF Neuron}}
Unlike the ALIF neuron,  the BRF neuron maintains a complex membrane potential. All operation counts are expressed in terms of real multiplications and real additions. As illustrated in Fig. 5, each complex multiplication is decomposed into four real multiplications and two real additions. Implementing the update \eqref{eq:rf_dynamics}, the block diagram in Fig.~\ref{fig:digital_brf} requires four multiplication and four addition operations. Furthermore, the smooth reset mechanism in \eqref{eq:dampening} introduces a further addition.  Fig.~\ref{fig:digital_brf}(b) shows the adaptive threshold update \eqref{eq:adap_vthr_brf}, which involves two additions and one multiplication. Overall, the number of operations for the soma of a BRF neuron consists of $N_{\rm add}^{\rm som}=6$ sums and $N_{\rm mul}^{\rm som}=5$ multiplications at each time step, as well as $N_{\mathrm{add}}^p=1$ sum and no multiplications, i.e., $N_{\rm mul}^p=0$, for each spike processed by the neuron. 

The input current is evaluated through synaptic operations as for ALIF neurons. The total number of synaptic additions thus equals the number of input spikes $\check{M}_{t,i}^l \leq n_i^l$, i.e., $N_{\rm add}^{\rm syn} = \check{M}_{t,i}^l$. 

The LIF neuron can be viewed as a simplified case of ALIF without the adaptive threshold mechanism, and its high-level architecture is well established in the literature \cite{10242251}. Furthermore, the RF neuron can be obtained as a special case of the BRF neuron by removing the adaptive threshold and refractory mechanism, and hence its architecture follows directly from Fig. \ref{fig:digital_brf}.

\subsubsection{Energy Consumption Model}
Denote by $E_{\rm add}$ the energy of addition operation and by $E_{\rm mul}$ the energy cost of multiplication. For example, for 45 nm CMOS technology, we have  $E_{\rm add}=0.1$ pJ for a 32-bit addition and  $E_{\rm mul}=3.2$ pJ for a 32-bit multiplication \cite{horowitz20141}. For a network layer $l$ with $K^l$ neurons, number of time steps $T$, and with a total number of input spikes across all $K^l$ neurons equal to $\check M_{t}^l$  and a total number of output spikes equal to $\hat M_{t}^l=\sum_{i=1}^{K^l} S_{t,i}^l$ in layer $l$, the energy consumption due to the somatic operations is given by 
\begin{align} \label{eq:energy_soma}
E^l_{\mathrm{soma}} = &~ T \cdot K^l \cdot (N^{\rm som}_{\mathrm{add}}E_{\rm add} + N^{\rm som}_{\rm mul}E_{\rm mul}) \notag \\
&~ + \sum_{t=1}^T \hat M_{t}^l(N^p_{\mathrm{add}}E_{\rm add} + N^p_{\rm mul}E_{\rm mul}). 
\end{align}
Furthermore, the energy consumed by synaptic operations is 
\begin{equation} \label{eq:energy_synapse}
E^l_{\rm synapse} = \sum_{t=1}^T \check M_{t}^lE_{\rm add},
\end{equation}
where $\check M_{t}^l = \sum_{i=1}^{K^l} \check M_{t,i}^l $ is the total number of synaptic operations for layer $l$ at time $t$.
Finally, the total energy consumption in layer $l$ is 
\begin{equation} \label{eq:energy_total}
E^l = E^l_{\rm synapse} + E^l_{\rm soma}.
\end{equation}

\section{Simulation Results} \label{sec:exp}

In this section, we present simulation results to investigate the advantages of our proposed neuromorphic split computing system using BRF neurons for processing inputs with rich frequency-domain features\footnote{Code is available at: https://github.com/kclip/neurocomm-rf }. We begin with audio signals, which consist of real-valued inputs, and then extend our analysis to complex-valued baseband signals. All experiments were carried out using PyTorch on a GPU server with a single NVIDIA A100 card.

\subsection{Settings}
We evaluate the performance on an event-based audio dataset, namely the SHD \cite{9311226}, and on one baseband signal classification benchmark, the ITS dataset \cite{girmay2023technology}. The choice of these tasks is informed by the intended application of the proposed split neuromorphic system, which includes the efficient processing of audio and communications signals. 

\subsubsection{\textit{Audio Processing}} The SHD dataset consists of 10,420 recordings of ten spoken digits (0 to 9) in English and German. To generate the dataset, the raw audio recordings were converted into spike trains using a biologically inspired model of the inner ear and auditory pathway \cite{9311226}. Following the methodology in \cite{yin2021accurate,higuchi2024balanced}, we work directly on the spike-encoded audio in the SHD dataset, discretizing the spike trains using time steps of $4\,$ms, resulting in input sequences of 250 time steps, with zero-padding as needed.

\subsubsection{\textit{Baseband Processing}} The ITS dataset \cite{girmay2023technology} includes six signal categories of complex-valued baseband signals -- LTE, 5G NR, Wi-Fi, C-V2X PC5, and ITS-G5 -- captured in the 5.9~GHz intelligent transportation systems (ITS) band. Each signal consists of in-phase and quadrature (IQ) samples at multiple sampling rates (1–25~Msps). In this work, we use the samples recorded at 5~Msps, since this choice was shown in \cite{girmay2023technology}  to guarantee near-optimal accuracy, while limiting the sampling rate. Each class contains 7,500 samples recorded over a fixed time window of 44~$\mu$s. 

\subsection{Architectures}
In terms of architectures, for both tasks, the RF and LIF configurations are obtained by removing the adaptive threshold mechanism from the BRF and ALIF neurons, respectively. Furthermore, to isolate the impact of smooth reset mechanism introduced by BRF, RF neurons use a soft reset mechanism. For split computing, the network is partitioned at the output of the second layer. 

\subsubsection{\textit{Audio Processing}} For the SHD dataset, we implement all schemes using a two-layer fully connected (FC) architecture, comparing the performance of architectures with LIF, ALIF, RF, and BRF neurons. Specifically, a 700-RFC128-RFC128-O20 architecture is used, where the input size is 700, RFC128 denotes an FC layer with 128 neurons and recurrent connections, and O20 corresponds to an output layer with 20 output classes. As discussed, the SHD dataset is generated by using a biologically inspired cochlear model that converts raw audio waveforms into spike trains through an auditory front-end.  Following standard practice \cite{higuchi2024balanced, yin2021accurate}, we feed the spike trains directly to SNN models to process the input data.

\subsubsection{\textit{Baseband Processing}} For the ITS dataset, in contrast to the FFT-based preprocessing used in \cite{girmay2023technology}, we adopt an architecture that employs a fully learnable configuration of 1-FC$^*$10-FC128-FC128-O6, with the complex-valued signal as input. The encoding layer, FC$^*$10, performs a complex linear transformation and produces 10 output neurons. Following~\cite{girmay2023technology}, the accuracy performance of all models is evaluated under 0~dB SNR with additive white Gaussian noise.

\subsection{Training}
We adopt the initialization strategy from \cite{yin2021accurate} for LIF-based SNNs, and follow the procedure outlined in \cite{higuchi2024balanced} for RF-based SNNs. For the SHD dataset, the network is trained for 20 epochs with a batch size of 32. For the ITS dataset, we train the network for 20 epochs using a batch size of 128. For each dataset, we use the same SNN architecture across all neuron types. Only the neuron model is varied, while the network topology, number of layers, and training procedure remain unchanged.

All neuron models are trained under the same initialization and training protocol to ensure a fair comparison, and we do not use a separate validation set, since our focus is on controlled comparisons rather than hyperparameter optimization. The learning rate follows a cosine annealing schedule starting from 0.075 and decaying to 0. Training is performed with backpropagation through time using surrogate gradients as described in \cite{higuchi2024balanced}, where the complex-valued BRF dynamics are implemented by separating real and imaginary components so that gradients are propagated through real-valued variables.

We apply the regularization term defined in \eqref{eq:regularizer} with sparsity coefficient $\alpha$ varying over a range from $1 \times 10^{-5}$ to $5 \times 10^{-3}$. For the ITS dataset, we vary the sparsity coefficient $\alpha$ over a range from $1 \times 10^{-5}$ to $9 \times 10^{-5}$ for all neuron models. These ranges were identified through an initial coarse search.

\subsection{Performance Metrics}
We evaluate the compute energy using \eqref{eq:energy_total}, which accounts for the somatic operation via \eqref{eq:energy_soma} and for the synaptic operations via \eqref{eq:energy_synapse}. Synaptic operations at the input layer are excluded from the compute energy estimation, as they introduce a constant overhead that is identical across models. Transmission energy per OFDM symbol is calculated using \eqref{tran_ene} multiplied by the OFDM symbol duration of 35.68~$\mu \mathrm{s}$, which is used in 5G NR \cite{NR}.

To obtain statistically reliable performance estimates, both accuracy and compute energy are evaluated using a bootstrap resampling procedure on the test dataset \cite{Japkowicz_Boukouvalas_2024}. Specifically, the test dataset is resampled with replacement 30 times, and model evaluation is repeated for each bootstrap replicate. The final reported accuracy and compute energy correspond to the median of the 30 runs, with the 5th–95th percentile range indicating a confidence interval.

\subsection{Channel Model}
We consider a five-path frequency-selective channel where each path amplitude follows a Rayleigh distribution with equal average power. The channel is normalized such that the average channel norm equals 1. For the path loss, we adopt the InH-Office Line-of-Sight (LOS) model from 3GPP TR 38.901 \cite{zhu20213gpp}. Accordingly, the path loss $P_L$ in dB is given by
\begin{equation}
    P_L=32.4+17.3 \log_{10}(d)+20\log_{10}(f_c), \label{path_loss}
\end{equation}
where $d$ is the distance between the Tx and Rx in meters, and $f_c$ is the carrier frequency, which is set to 6G Hz.  The channel SNR is defined as the ratio of the per-subcarrier transmission power $P$ over the noise power, scaled by the path loss, i.e.,  
\begin{equation}
    \text{SNR}= \frac{P}{N_0} \cdot 10^{-\frac{P_L}{10}}. 
\end{equation}
If not stated otherwise, the average SNR is set to 20 dB. 
 
\subsection{Audio Processing: Results}
We start with results for the audio processing task.
\subsubsection{Centralized Benchmark}
To establish a baseline, we first assess the performance of the four spiking neuron models -- LIF, ALIF, RF, and BRF -- in a centralized setting with a single, undivided SNN.

\begin{figure}[ht!]
    \centering
    \includegraphics[width=1.0\columnwidth]{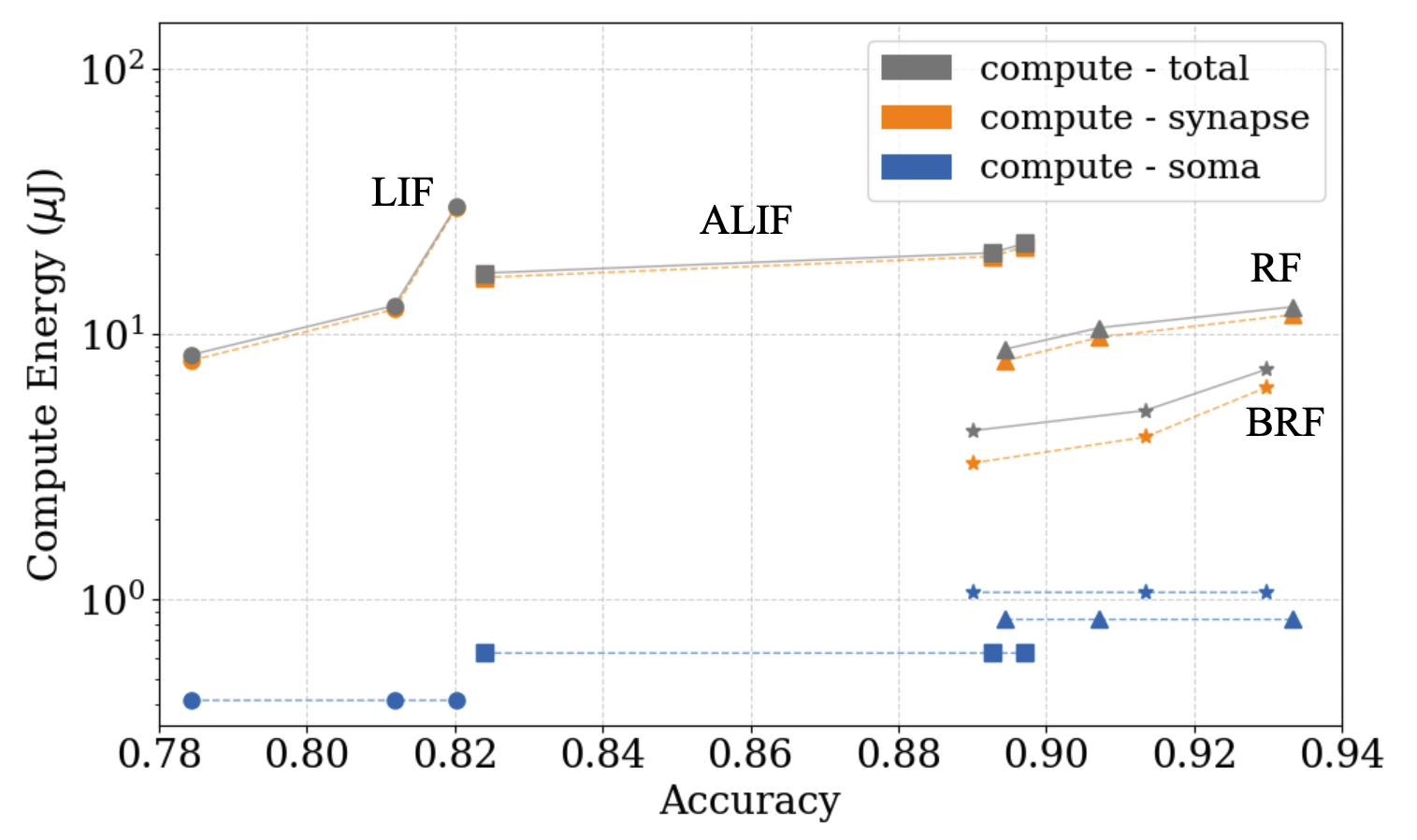}
    \caption{Energy consumption versus accuracy for different neuron models on the SHD dataset. The compute energy consists of somatic and synaptic components, and curves are obtained by varying the sparsity level via the regularization coefficient $\alpha$ in \eqref{eq:l_reg}. The confidence intervals are omitted to improve visual clarity.
    }
    \label{fig:shd_sparsity}
\end{figure}

Fig.~\ref{fig:shd_sparsity} illustrates the compute energy, including the separate contributions of synaptic and somatic operations, as a function of the achieved classification test accuracy. The curves are obtained by varying the sparsity level via the regularization coefficient $\alpha$ in \eqref{eq:l_reg}. Both RF and BRF models leverage the inherent resonant property of their internal dynamics to enhance the sparsity of their synaptic activity by selectively responding to temporally structured inputs. As a result, both RF and BRF neurons significantly reduce the compute energy, while also achieving a higher accuracy. 
For example, the SNN with RF neurons achieves an accuracy up to 93.3\% with energy consumption as low as 8.77~$\mu\mathrm{J}$. 
Since soma operations involve only a fixed and relatively small number of additions and multiplications per time step, while synaptic energy scales with every incoming spike, the contribution of soma energy remains consistently minor compared to synaptic energy across all neuron types.

By incorporating an adaptive threshold mechanism, BRF neurons can further reduce energy usage to 4.33~$\mu\mathrm{J}$, while maintaining comparable accuracy. In contrast, the simpler ALIF and LIF models yield lower accuracy—up to 89.7\% for ALIF and 82.0\% for LIF—and significantly higher energy consumption, i.e., 22.01~$\mu\mathrm{J}$ and 30.44~$\mu\mathrm{J}$, respectively. Note that for LIF models with accuracy levels near 82\%, the significant jump in energy implies that unregularized LIF neurons have a significant number of redundant spikes. These results also indicate that the somatic operations introduced by the more complex BRF neuron entail negligible additional energy, as the synaptic operations dominate the overall energy consumption. Furthermore, these results confirm that the intrinsic frequency selectivity of RF and BRF neurons enables them to achieve higher spectral discrimination, directly translating to improved accuracy–energy efficiency compared to conventional LIF and ALIF models.

To explicitly present the confidence intervals omitted in Fig. 6 for visual clarity, Table I in Appendix provides the full statistical results, including the median, 5th percentile, and 95th percentile for all operating points.

\subsubsection{Split Computing Architecture} We now evaluate the performance of the pre-trained models from Fig.~\ref{fig:shd_sparsity} in a split computing setting under the channel model described above. We specifically consider SNNs with ALIF, RF, and BRF neuron models that achieve accuracy levels close to 90.0\% in the centralized setting of Fig.~\ref{fig:shd_sparsity}. This choice excludes SNNs with LIF neurons due to their lower accuracy. The transmission power $P$ in \eqref{tran_ene} is dynamically adjusted to maintain a fixed average SNR equal to 20~dB per subcarrier at the receiver.

\begin{figure}[ht!]
    \centering
    \includegraphics[width=1.0\columnwidth]{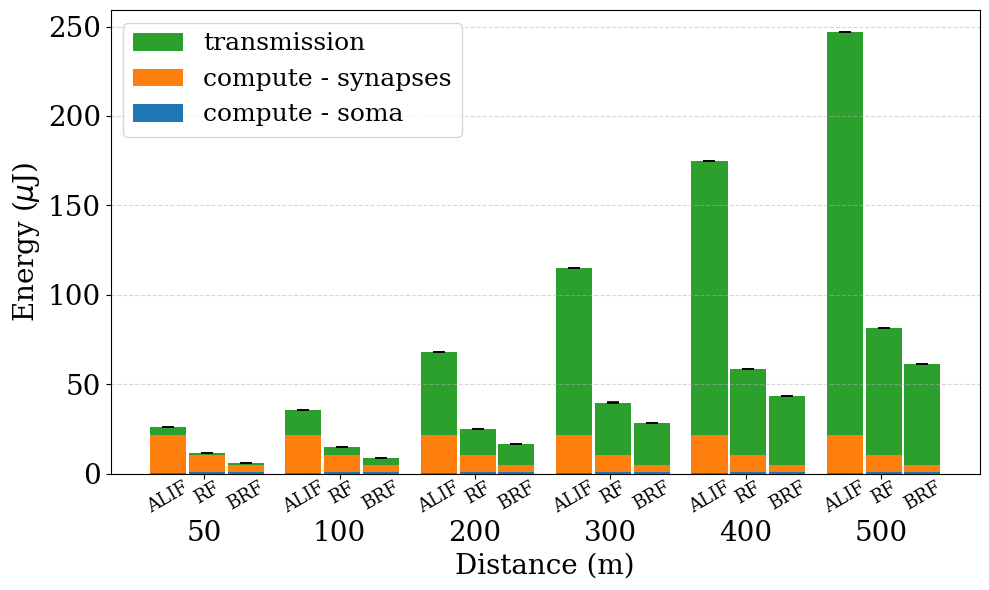}
    \caption{Energy consumption versus communication distances with adaptive transmission energy allocation on the SHD dataset. Each bar shows the breakdown into somatic (blue), synaptic (orange), and transmission (green) energy contributions. The confidence intervals are extremely narrow, appearing as thin lines on the bars.}
    \label{fig:shd_adaptive_energy}
\end{figure}

Fig.~\ref{fig:shd_adaptive_energy} presents the breakdown of the total energy consumption into somatic, synaptic, and transmission components across a range of communication distances. As shown in the figure, at short communication ranges, e.g., no larger than 50 $\rm m$, the compute energy dominates the total energy consumption. For instance, at 50 $\rm m$, the BRF model incurs approximately $5.16~\mu \rm J$ in compute energy, while the transmission energy remains lower at $1.05~\mu \rm J$. As the communication distance increases, the transmission energy becomes the dominant factor, scaling approximately quadratically with distance due to free-space path loss in \eqref{path_loss}. At $100~\mathrm{m}$, the BRF's transmission energy rises to $3.47~\mu \rm J$, and at $200~\mathrm{m}$, it increases to $11.53~\mu \rm J$.

\begin{figure}[t!]
    \centering
    \includegraphics[width=1.0\columnwidth]{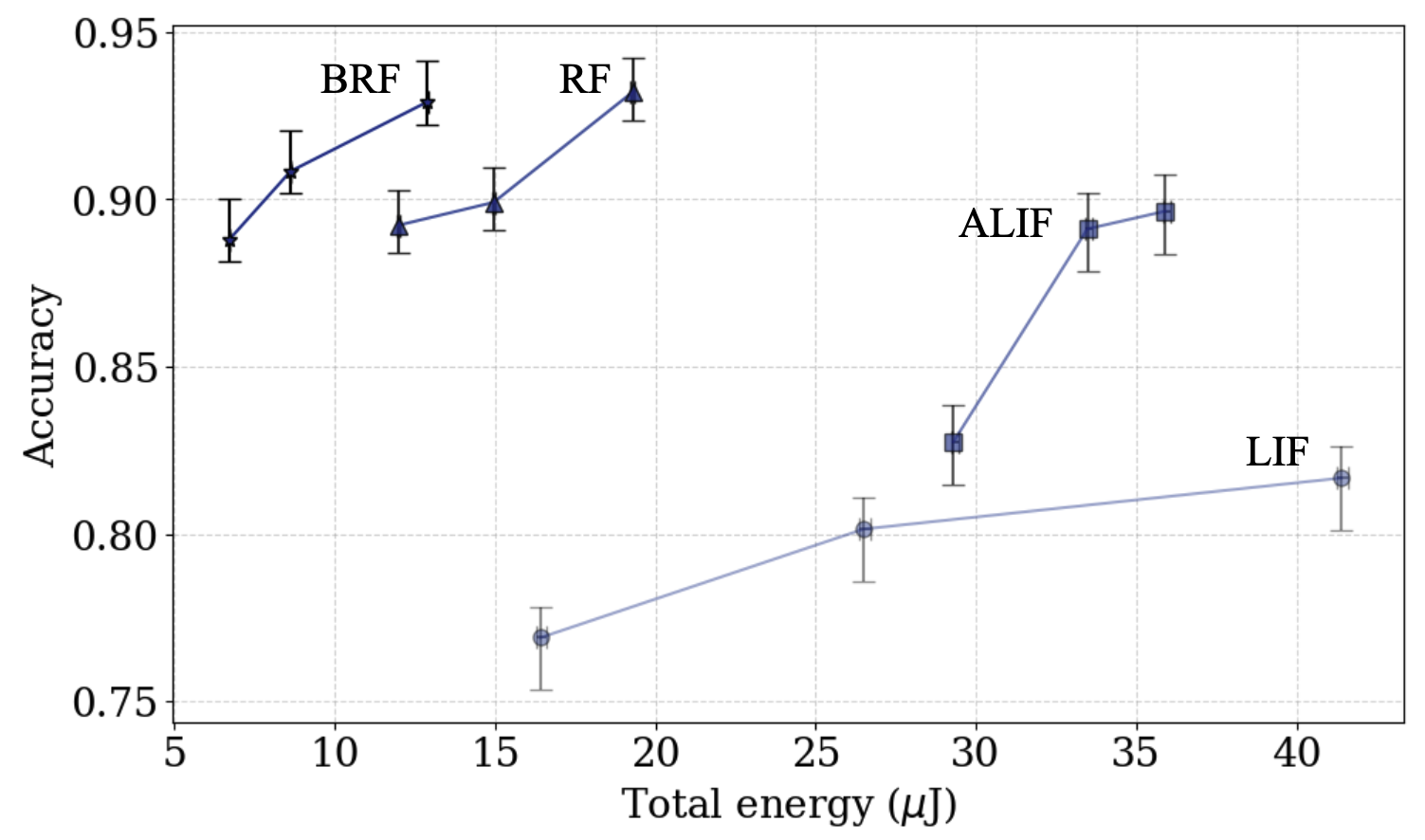}
    \caption{Accuracy versus total energy consumption, encompassing both compute and transmission, at the communication distance of 100 $\rm m$ on the SHD dataset. For each model, the curves reflect varying regularization strengths  $\alpha$ in \eqref{eq:l_reg}. Most confidence intervals for total energy are extremely narrow and therefore not visible in the figure. }
    \label{fig:shd_total_energy}
\end{figure}

Fig.~\ref{fig:shd_total_energy} shows the accuracy versus the total energy consumption, including both compute and transmission energy. As in Fig.~\ref{fig:shd_sparsity}, each point in the figure corresponds to a model trained under a different regularization strength $\alpha$. The BRF model achieves the best energy-accuracy trade-off, reaching up to 92.9\% accuracy with total energy of as low as 6.73~$\mu \rm J$. RF achieves similar accuracy, but at the higher energy cost of 11.99~$\mu \rm J$. ALIF reaches up to 89.6\% accuracy but requires 29.30 to 35.91~$\mu \rm J$, while LIF reaches only 81.7\% accuracy at an energy cost as high as 41.39~$\mu \rm J$. Overall, despite the BRF neuron's increased somatic complexity, the associated energy cost is negligible, and the substantial spike sparsity levels lead to lower synaptic and transmission energy consumption, resulting in the most efficient energy-accuracy trade-off.

\subsection{Baseband Processing: Results}
We now turn to results for the ITS dataset. 
\subsubsection{Centralized Benchmark} 

\begin{figure}[!ht]
    \centering
    \includegraphics[width=1.0\columnwidth]{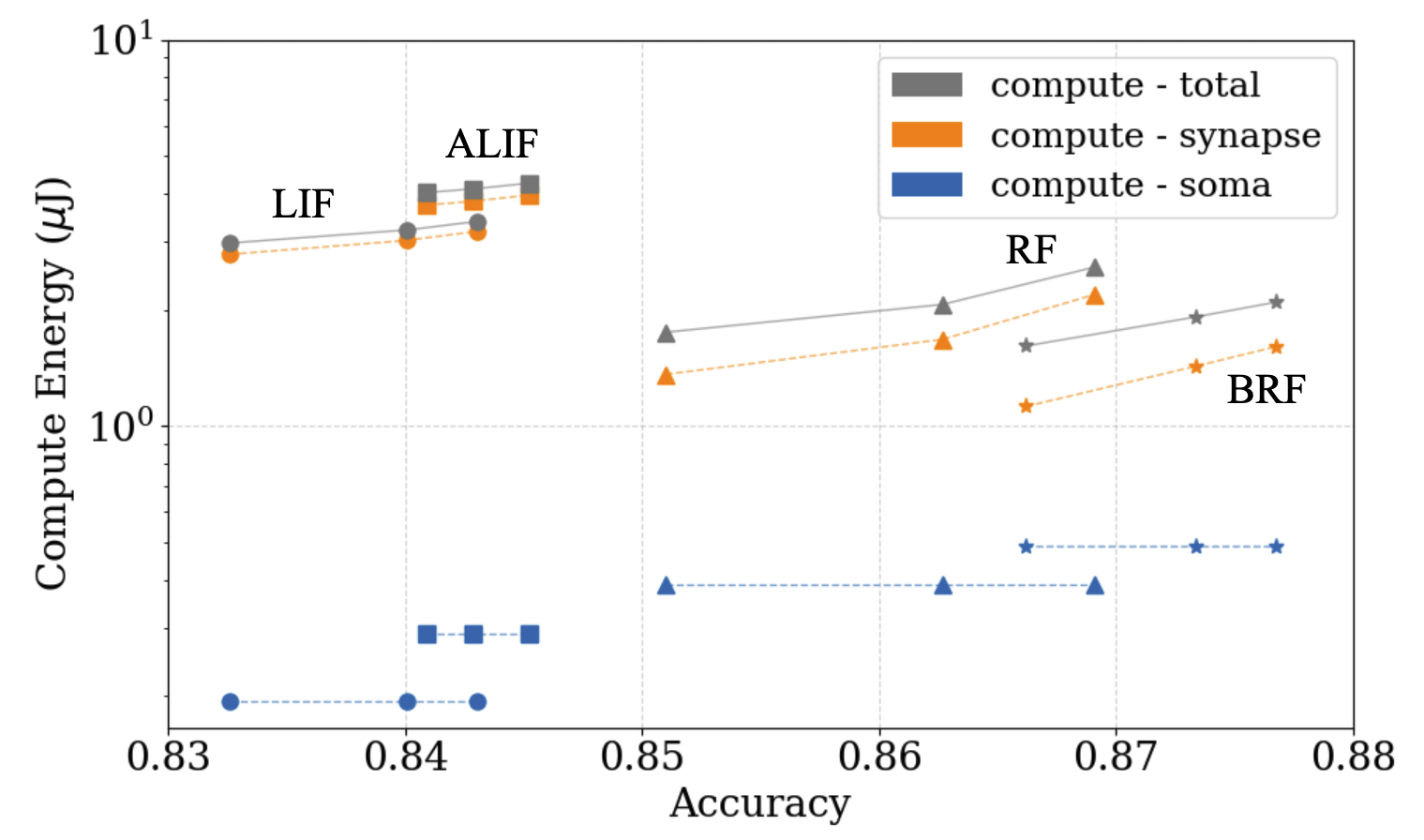} 
    \caption{Energy consumption versus accuracy for different neuron models on the ITS dataset. The compute energy consists of somatic and synaptic components, and curves are obtained by varying the sparsity level via the regularization coefficient $\alpha$ in \eqref{eq:l_reg}. The confidence intervals are omitted in this figure for clarity.} 
    \label{fig:its_sparsity}
\end{figure}

Consider first a centralized implementation. As shown in Fig.~\ref{fig:its_sparsity}, in a manner similar to Fig.~\ref{fig:shd_sparsity}, BRF and RF models demonstrate superior energy-accuracy trade-offs compared to ALIF and LIF.  BRF and RF neurons provide advantages in both energy and accuracy compared to LIF and ALIF, with the BRF model achieving the highest accuracy at approximately 87.7\%, with compute energy as low as 1.61~$\mu \rm J$. The RF model performs comparably, reaching 86.9\% accuracy with compute energy around 1.75~$\mu \rm J$. In contrast, ALIF and LIF yield lower accuracy—around 85.0\% and 84.0\%, respectively—while consuming significantly more energy, i.e., up to  4.26~$\mu \rm J$ for ALIF. Confidence intervals, not shown for visual clarity, are generally consistent with those in Fig. \ref{fig:shd_sparsity} (see Appendix).

\begin{figure}[ht!]
    \centering
    \includegraphics[width=1.0\columnwidth]{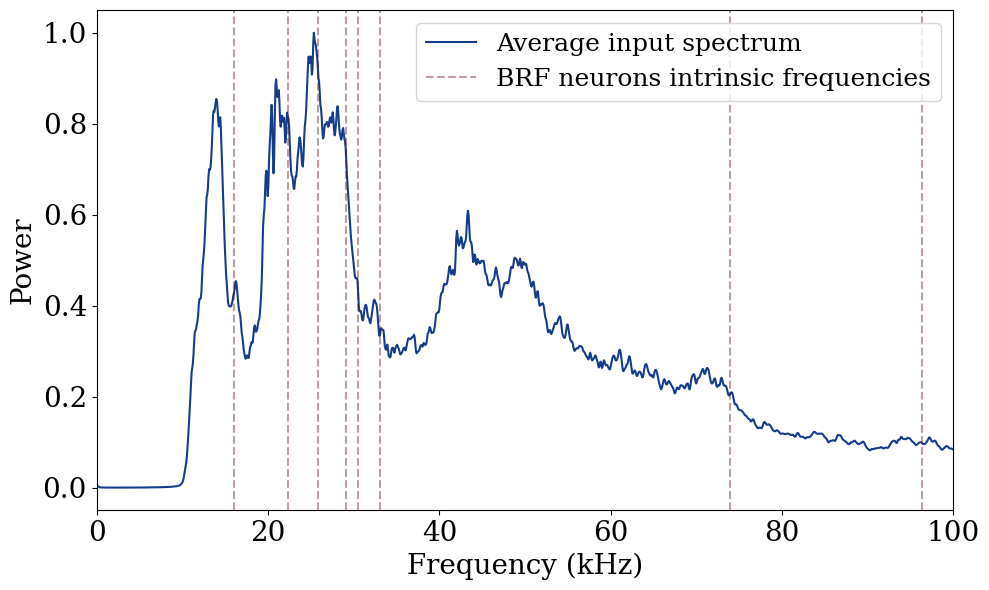}
    \caption{Illustration of the average spectrum of ITS baseband signals (solid line) and the intrinsic frequencies (vertical dashed lines) of BRF neurons in the encoding layer.}
    \label{fig:its_frequency}
\end{figure}

An explanation for the gains of RBF neurons can be obtained by comparing the spectrum of the input baseband signals to the intrinsic frequencies learned by the encoding layer of RBF neurons. To elaborate on this, Fig.~\ref{fig:its_frequency} shows the squared absolute value of the Fourier transform of the baseband signals, averaged over test examples, along with vertical lines denoting the intrinsic frequencies (measured in Hz) for the trained encoding RBF neurons. It is observed that the RBF neurons operate at intrinsic frequencies that match well with the parts of the spectrum at which the input signals contain the most energy. This confirms that the BRF neurons are capable of extracting informative spectral features from the input signals.

Furthermore, we note that multiple neurons often concentrate around high-energy regions of the spectrum. In contrast to standard designs tailored for signal analysis and synthesis, this clustering is not redundant. Rather, it reflects task-driven specialization for the task of classification. In fact, overlapping frequency selectivity can increase robustness in the decision-making process emphasizing task-relevant spectral features.

\subsubsection{Split Computing Architecture}

We now examine the performance of the pre-trained BRF and RF models from Fig.~\ref{fig:its_sparsity} under a wireless split computing setup. These two models are selected for their strong energy-accuracy performance, achieving accuracy levels near 87.0\% in the centralized setting.

\begin{figure}[!ht]
    \centering
    \includegraphics[width=1.0\columnwidth]{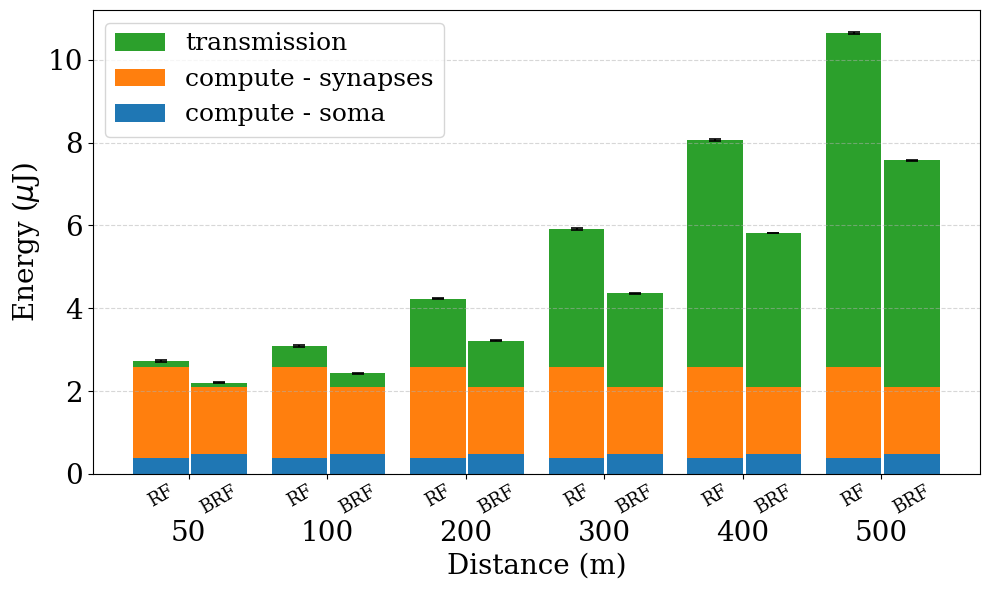}
    \caption{Energy consumption versus communication distances with adaptive transmission energy allocation on the ITS dataset. Each bar shows the breakdown into somatic (blue), synaptic (orange), and transmission (green) energy contributions. The confidence intervals are extremely narrow, appearing as thin lines on the bars.}
    \label{fig:its_adaptive_energy}
\end{figure}

Fig.~\ref{fig:its_adaptive_energy} illustrates the breakdown of total energy consumption into somatic, synaptic, and transmission components across varying communication distances for both BRF and RF models. In a manner similar to Fig.~\ref{fig:shd_adaptive_energy}, at short distances, compute energy dominates the total energy budget. For instance, at 50~$\mathrm{m}$, BRF and RF consume 2.09~$\mu\mathrm{J}$ and 2.58~$\mu\mathrm{J}$ in compute energy, respectively, while their corresponding transmission energy remains relatively low at 0.10~$\mu\mathrm{J}$ for BRF and 0.15~$\mu\mathrm{J}$ for RF.
As the communication distance increases, transmission energy becomes the dominant contributor to total energy. The transition point, where transmission energy surpasses compute energy, occurs between 200 and 300~$\mathrm{m}$ for both models. For example, at 300~$\mathrm{m}$, transmission dominates for RF, rising to 3.33~$\mu\mathrm{J}$.

\begin{figure}[h!]
    \centering
    \includegraphics[width=1.0\columnwidth]{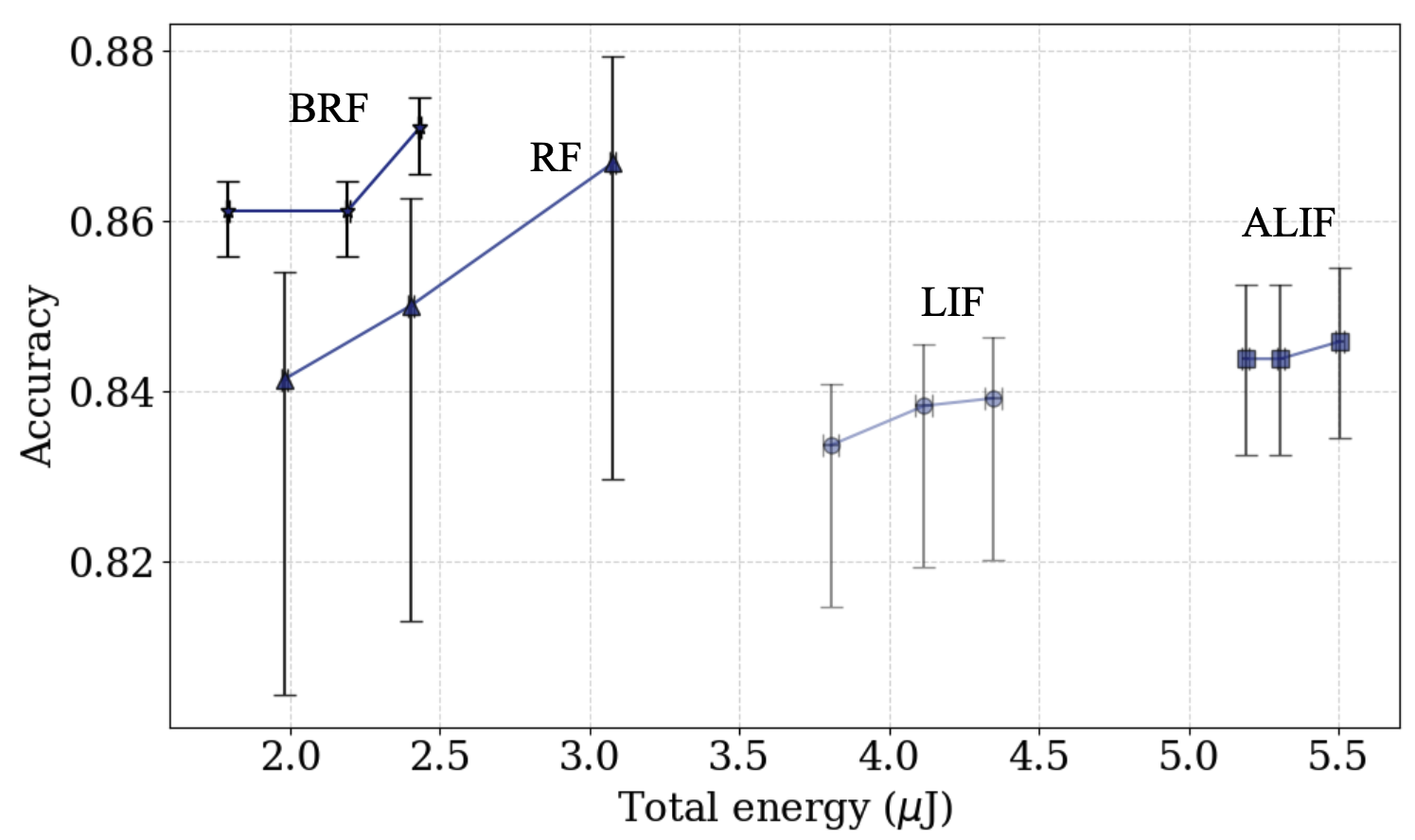}
    \caption{Accuracy versus total energy consumption, encompassing both compute and transmission, at the communication distance of 100 $\rm m$ on the ITS dataset. For each model, the curves reflect varying regularization strengths  $\alpha$ in \eqref{eq:l_reg}. The confidence intervals for total energy are small, indicating small variance.}
    \label{fig:its_total_energy}
\end{figure}

Fig.~\ref{fig:its_total_energy} presents the accuracy versus total energy consumption at a communication distance of 100~$\mathrm{m}$, with each point corresponding to a different regularization strength $\alpha$. The BRF and RF models exhibit similar accuracy performance, achieving up to 87.1\% accuracy. However, BRF consistently requires less energy, with total consumption as low as 1.80~$\mu\mathrm{J}$ compared to 2.45~$\mu\mathrm{J}$ for RF. In contrast, the ALIF and LIF models require significantly more energy—up to 5.51~$\mu\mathrm{J}$ and 4.38~$\mu\mathrm{J}$, 
respectively—while reaching lower accuracy levels roughly 84.0\% to 85.0\%. These results are consistent with the conclusions drawn for the SHD dataset, where the BRF neuron were seen to achieve the most efficient energy-accuracy trade-off. The high variance observed in Fig. \ref{fig:its_total_energy} can be attributed to the fact that the models are trained on clean data but evaluated by adding 0 dB white noise to the clean input. However, models incorporating BRF neurons exhibit more stable predictions than those employing other neuron types.

\subsection{Quantization and Calibration: Results}

\begin{figure}[!ht]
    \centering
    \includegraphics[width=1.0\columnwidth]{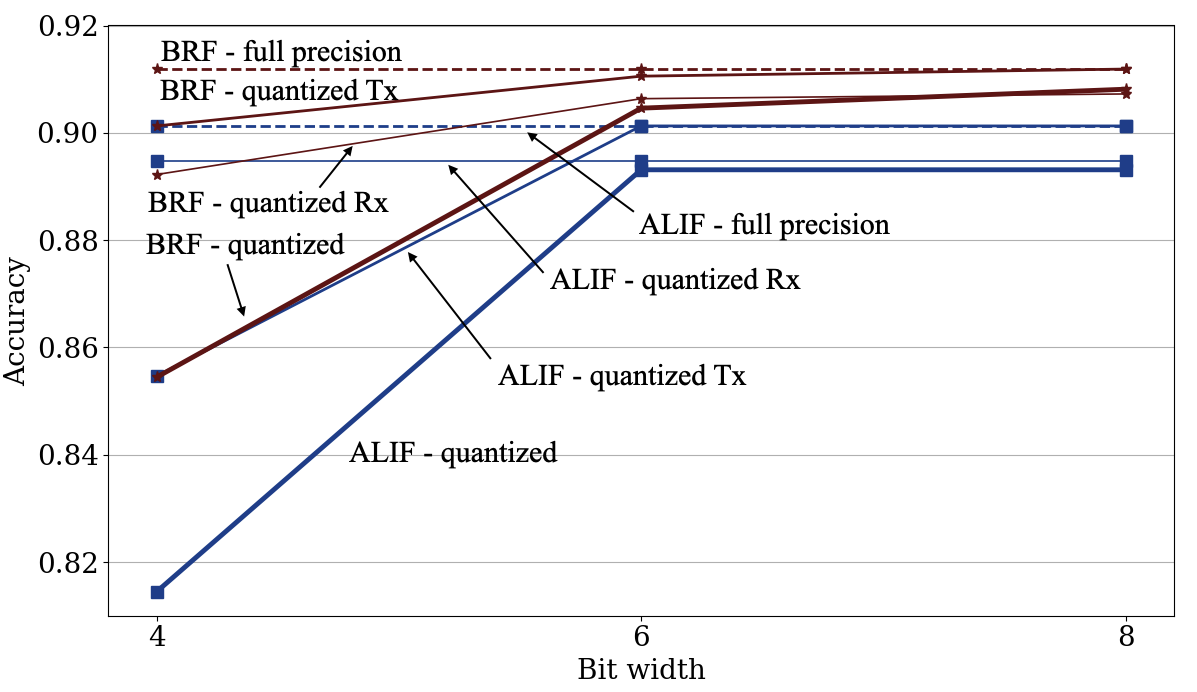}
    \caption{Accuracy versus bit width for quantized models with ALIF and BRF neurons on the SHD dataset. The comparison includes fully quantized models, as well as settings where only the transmitter or receiver is quantized. The dashed line indicates the full-precision baseline. All models are calibrated using  \eqref{eq:calibration} with 2\% of the training data.}
    \label{fig:shd_quantization}
\end{figure}

So far, we have considered full-precision models. We now illustrate the accuracy of different models after quantization. Fig.~\ref{fig:shd_quantization} shows classification accuracy for the SHD dataset when the number of bits per weight varies from 4 to 8. BRF models are seen to exhibit stronger resilience to quantization than ALIF models across all bit widths. Calibration is performed using 2\% of the training samples by following the approach described in Sec.~\ref{sec:quantization}. We consider a fully quantized model, as well as settings in which only the transmitter model or the receiver model is quantized. These scenarios may model cellular communication setups in which one end of the communication link, namely the base station, has less stringent computational limitations. 

For 4-bit models with calibration, the fully quantized ALIF model improves from 52.1\% to 81.4\%, while the BRF model improves from 80.6\% to 85.4\%. When only the transmitter is quantized, the calibrated accuracy reaches 85.5\% for the ALIF model and 90.1\% for the BRF model, indicating that ALIF is more sensitive to reduced precision in the transmitter-side encoder. In contrast, quantizing only the receiver results in minimal degradation for both neuron models: the ALIF model improves from 88.7\% to 89.5\%, and the BRF model from 89.0\% to 89.2\% after calibration.

We further observe that 4-bit receiver-only quantization leads to a smaller accuracy drop for ALIF (0.7\%) compared to BRF (2.0\%) relative to their full-precision baselines. This suggests that ALIF relies more heavily on precise encoding at the transmitter, allowing the receiver to operate at low bit precision. Conversely, the slightly larger accuracy drop for the BRF model indicates that its performance is more uniformly affected by both transmitter and receiver quantization. Given the sparse and highly temporal nature of spike patterns in the BRF model, this reflects that accurate decoding in the BRF receiver requires high bit precision to match full-precision performance.

Overall, these considerations support the results shown in Fig.~\ref{fig:shd_quantization}: the BRF model achieves higher accuracy under quantization, while the performance of the ALIF model hinges on the preservation of the level in the early encoding layers. The higher resilience of the BRF neurons to quantization may be due to their capacity to filter out noise that does not align with their respective resonant frequencies.

\subsection{Impact of Pre-Processing}
In the previous experiments, LIF- and ALIF-based SNNs were fed directly the input signals. While this is a natural choice for neuromorphic data, as for the SHD dataset, in the case of natural signals one could pre-process the inputs using a standard FFT. We now compare the performance of LIF, ALIF, RF, and BRF solutions, as studied so far, with LIF-based SNNs or ANNs that are preceded by an FFT.

All models share the same architecture for processing features encoded from complex-valued signals, consisting of two fully connected layers with 128 neurons each. The FFT produces feature vectors of sizes 16, 64, or 128. The energy consumption of FFT processing is estimated using the digital Radix-2 implementation ~\cite{mookherjee2015low}, which reports the number of addition and multiplication operations required for each FFT configuration. For example, for an input length of 256, the computation consumes approximately 0.005 $\mu \rm J$ per input sample, considering 45 nm CMOS technology \cite{yu2012design}.

\begin{figure}[!h]
    \centering
    \includegraphics[width=1.0\linewidth]{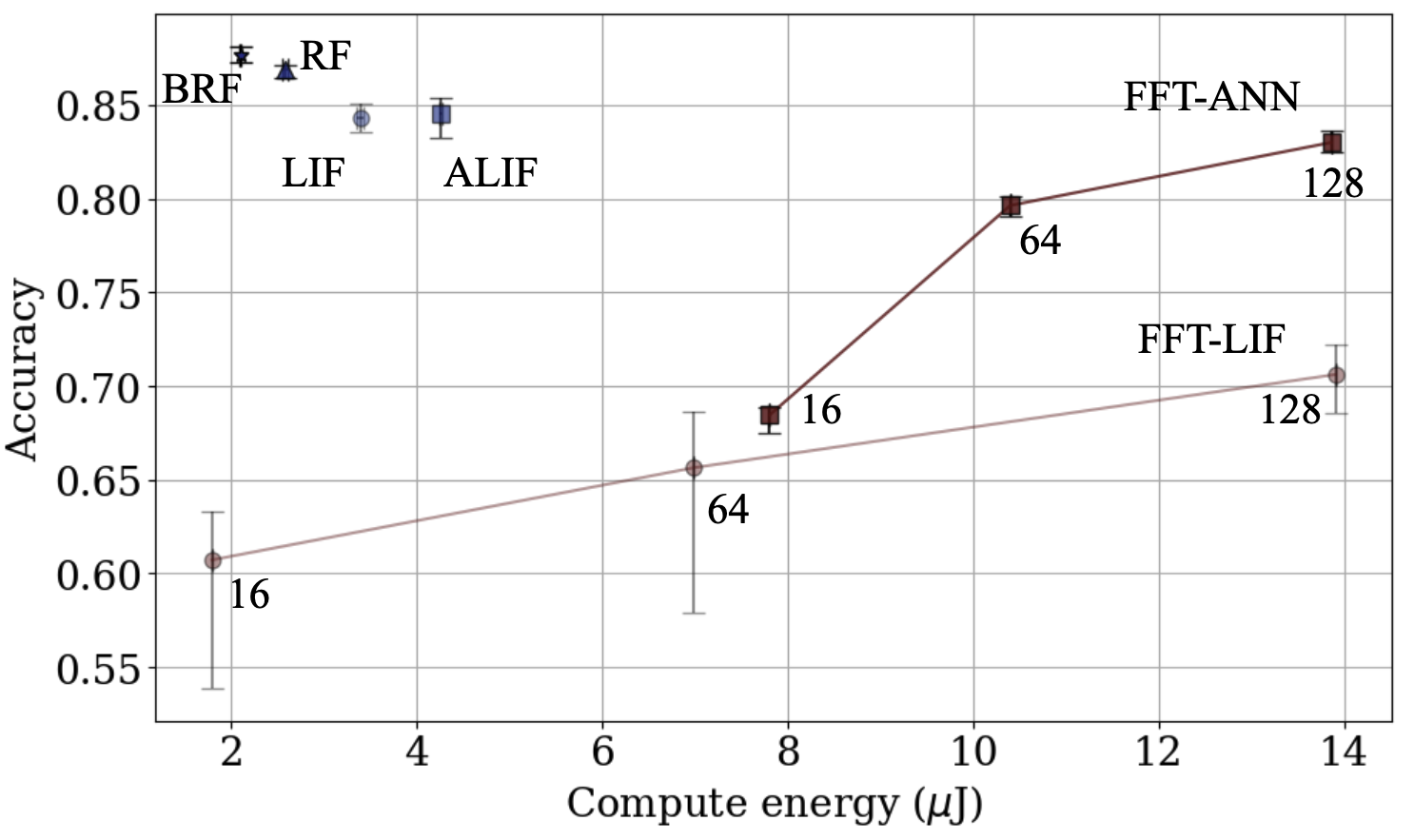}
    \caption{Accuracy versus compute energy for non-FFT models using LIF, ALIF, RF and BRF neurons and FFT-based models. The FFT-based models (FFT-ANN and FFT-SNN) use FFT pre-processing with 16, 64, or 128 linearly spaced frequency components.}
    \label{fig:fft_noiseless}
\end{figure}

Fig. \ref{fig:fft_noiseless} compares the accuracy and compute energy of the mentioned solutions. In the FFT-based models, most of the computational energy is consumed by the matrix multiplication in the first layer. With 128 FFT features, the FFT-ANN exhibits approximately $7\times$ higher energy consumption than the BRF model. Although reducing the number of filters lowers the energy consumption, it also leads to a degradation in accuracy. In the FFT-LIF model, the first layer acts as a spike encoder that converts the FFT features into spike trains over two time steps, resulting in higher energy consumption than the proposed models due to the matrix multiplication used in the encoding process. The lower accuracy of the FFT-LIF model confirms that the advantage of the proposed models is attributed to their ability to optimize end-to-end temporal dynamics, whereas the FFT-LIF model relies on an additional, less efficient encoding of pre-processed FFT features.

\begin{table*}[!t]
\centering
\caption{
Energy consumption and accuracy for Fig.~6 on the SHD dataset: median (subscript 50) and 5th and 95th percentiles (subscripts 5 and 95) for accuracy, as well as synaptic and total compute energy (in $\mu$J).
}
\label{tab:fig6_energy}
\scriptsize
\resizebox{\textwidth}{!}{%
\begin{tabular}{llccccccccccc}
\toprule
Model & Point &
Acc$_5$ & Acc$_{50}$ & Acc$_{95}$ &
$E_{\text{soma}}$ &
$E^{\text{syn}}_5$ & $E^{\text{syn}}_{50}$ & $E^{\text{syn}}_{95}$ &
$E^{\text{tot}}_5$ & $E^{\text{tot}}_{50}$ & $E^{\text{tot}}_{95}$ \\
\midrule
LIF  & left   & 0.771 & 0.784 & 0.794 & 0.416 &  7.870 &  7.961 &  8.138 &  8.286 &  8.377 &  8.554 \\
LIF  & middle & 0.801 & 0.812 & 0.826 & 0.416 & 12.351 & 12.409 & 12.447 & 12.767 & 12.825 & 12.863 \\
LIF  & right  & 0.805 & 0.820 & 0.834 & 0.416 & 29.901 & 30.024 & 30.179 & 30.317 & 30.440 & 30.595 \\
\midrule
ALIF & left   & 0.814 & 0.824 & 0.837 & 0.627 & 16.312 & 16.376 & 16.506 & 16.939 & 17.003 & 17.134 \\
ALIF & middle & 0.883 & 0.893 & 0.900 & 0.627 & 19.580 & 19.634 & 19.665 & 20.208 & 20.261 & 20.293 \\
ALIF & right  & 0.884 & 0.897 & 0.902 & 0.627 & 21.344 & 21.387 & 21.417 & 21.971 & 22.014 & 22.045 \\
\midrule
RF   & left   & 0.886 & 0.894 & 0.903 & 0.845 &  7.890 &  7.938 &  7.982 &  8.734 &  8.782 &  8.826 \\
RF   & middle & 0.897 & 0.907 & 0.919 & 0.845 &  9.687 &  9.737 &  9.777 & 10.531 & 10.582 & 10.622 \\
RF   & right  & 0.924 & 0.933 & 0.941 & 0.845 & 11.791 & 11.837 & 11.890 & 12.635 & 12.682 & 12.735 \\
\midrule
BRF  & left   & 0.882 & 0.890 & 0.902 & 1.062 &  3.250 &  3.265 &  3.279 &  4.312 &  4.327 &  4.341 \\
BRF  & middle & 0.901 & 0.913 & 0.923 & 1.062 &  4.077 &  4.094 &  4.110 &  5.140 &  5.157 &  5.173 \\
BRF  & right  & 0.919 & 0.930 & 0.940 & 1.062 &  6.273 &  6.299 &  6.320 &  7.335 &  7.361 &  7.382 \\
\bottomrule
\end{tabular}%
}
\label{table}
\end{table*}

\section{Conclusions} \label{con}
In this paper, we have presented a neuromorphic wireless split computing architecture that employs RF neurons for real-time time-series signal processing. Unlike conventional LIF neurons, RF neurons exhibit oscillatory dynamics that enable direct extraction of time-localized spectral features from streaming inputs, thereby eliminating the need for explicit spectral pre-processing. This results in lower spiking activity and, consequently, reduced computation and transmission energy. To enable wireless transmission of spike events, we developed an OFDM-based analog interface that efficiently maps neuronal spikes onto subcarriers, exploiting spike sparsity to minimize transmission overhead. Experimental evaluations on both audio and baseband radio-frequency signal classification tasks demonstrate that the proposed RF-SNN architecture achieves competitive accuracy compared to conventional LIF-SNNs and ANNs, while yielding substantial reductions in spike rates and overall energy consumption. These findings underscore the potential of RF-based neuromorphic models in enabling efficient, low-power intelligence at the wireless edge.    

Future work may extend the proposed wireless split SNN architecture with RF neurons to more complex sensory modalities, and investigate adaptive mechanisms for frequency tuning in dynamic environments. Another interesting direction for future research is to implement BRF neuron dynamics using analog circuits, which would enable direct circuit-level energy comparisons with recent neuromorphic RF designs \cite{11006495}.

\section*{Appendix}
In this Appendix, we report numerical results for Fig.~\ref{fig:shd_sparsity} in Table \ref{table}. The results correspond to the three points per curve represented in Fig.~\ref{fig:shd_sparsity}. More fine-grained results for Fig. 7, Fig. 8, Fig. 9, Fig. 11, Fig. 12, and Fig. 14 can be found in the supplementary material.

\small{
\bibliographystyle{IEEEtran}
\bibliography{references}}

\end{document}